\documentclass[10pt,twocolumn,letterpaper]{article}

\usepackage{wacv}              

\usepackage{graphicx}
\usepackage{amsmath}
\usepackage{amssymb}
\usepackage{booktabs}
\usepackage{makecell}
\usepackage{tabularx}
\usepackage[dvipsnames]{xcolor}
\usepackage[inline]{enumitem}

\newcommand{\mypar}[1]{\vspace{2mm}\noindent\textbf{#1}}
\newcommand{\ours}{DeCLIP\xspace}

\usepackage[pagebackref,breaklinks,colorlinks]{hyperref}

\usepackage[capitalize]{cleveref}
\crefname{section}{Sec.}{Secs.}
\Crefname{section}{Section}{Sections}
\Crefname{table}{Table}{Tables}
\crefname{table}{Tab.}{Tabs.}

\def\wacvPaperID{471} 
\def\confName{WACV}
\def\confYear{2025}

\begin{document}

\title{DeCLIP: Decoding CLIP representations for deepfake localization}

\author{Stefan Smeu\\
Bitdefender\\
{\tt\small ssmeu@bitdefender.com}
\and
Elisabeta Oneata\\
Bitdefender\\
{\tt\small eoneata@bitdefender.com}
\and
Dan Oneata\\
\textsc{Politehnica} Bucharest\\
{\tt\small dan.oneata@gmail.com}
}

\maketitle

\begin{abstract}
   Generative models can create entirely new images,
   but they can also partially modify real images in ways that are undetectable to the human eye.
   In this paper, we address the challenge of automatically detecting such local manipulations.
   One of the most pressing problems in deepfake detection remains the ability of models to generalize to different classes of generators.
   In the case of fully manipulated images,
   representations extracted from large self-supervised models (such as CLIP) provide a promising direction towards more robust detectors.
   Here, we introduce \ours---a first attempt to leverage such large pretrained features for detecting local manipulations.
   We show that, when combined with a reasonably large convolutional decoder, pretrained self-supervised representations
   are able to perform localization and improve generalization capabilities over existing methods.
   Unlike previous work, our approach is able to perform localization on the challenging case of latent diffusion models,
   where the entire image is affected by the fingerprint of the generator.
   Moreover, we observe that this type of data, which combines local semantic information with a global fingerprint, provides more stable generalization than other categories of generative methods.
   %
\end{abstract}

\begin{figure*}
    \centering
    \includegraphics[width=0.95\textwidth]{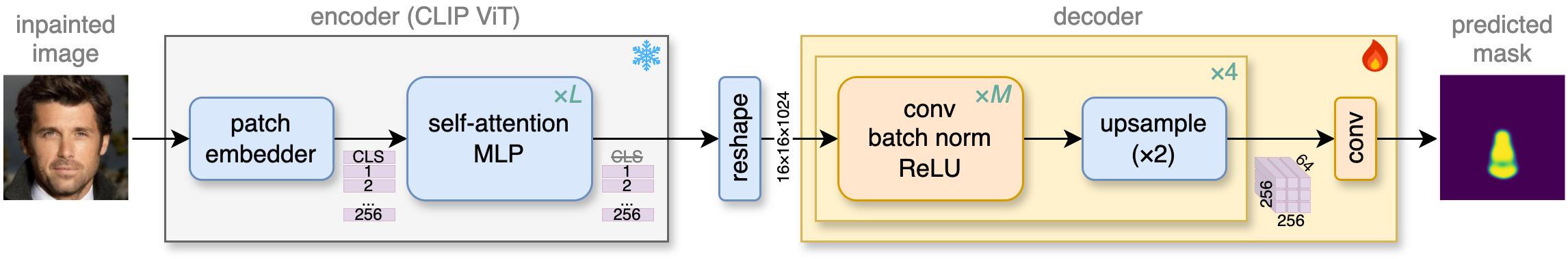}
\caption{%
    Method overview.
    We perform manipulation localization by decoding the information from the frozen CLIP embeddings using a learnt convolutional decoder.
    The embeddings are extracted at an arbitrary layer $L$ and upsampled progressively by the decoder.
}
\label{fig:overview}
\end{figure*}
\section{Introduction}
\label{sec:intro}

This paper addresses the task of localizing manipulations in partially altered images.
For example, given a video of a political figure whose mouth has been manipulated to make it look like they are uttering a certain sentence, we want to automatically identify this region as fake.
This type of manipulation, where most of the context is real and only a small part is manipulated,
is both highly deceptive because of the real context,
and easy to achieve because of the wide availability of inpainting techniques.
Precise localization of partial manipulations prevents this common type of attack and
provides a richer and more interpretable output than detection methods, which output a binary label (fake or real).

The main challenge of deepfake localization (and deepfake detection in general) remains the ability to generalize.
When training and test data are generated by similar methods, detection is possible \cite{Wang_2020_CVPR},
but when test data is generated by unseen methods, performance drops sharply \cite{lin2024detecting,liu2024evolving,ivanovska2024wacvw}.
Deepfake detectors, which are typically high-capacity networks,
rely on \emph{fingerprints} \cite{marra2019gans,Yu_2019_ICCV}---imperceptible patterns left by the generator.
But these fingerprints are sensitive to the generator (type \cite{corvi2022detection,ricker24visapp}, training data \cite{marra2019gans,Yu_2019_ICCV}, seed \cite{Yu_2019_ICCV}) hindering out-of-domain performance.
%
%
Recently, it has been shown that is possible to replace the very flexible detectors with representations produced by self-supervised models.
Specifically, Ojha \etal \cite{Ojha2023TowardsUF} extract features from the pretrained CLIP model \cite{clip} and use a linear classifier on top to distinguish fake from real images.
This simple approach shows strong generalization across a wide range of generators.
However, this method was only applied to fully manipulated images and used to predict image-level labels.

Our idea is to exploit the intrinsic generalization capability of CLIP features for the localization task.
To this end, we fill the gap in the literature by first evaluating these self-supervised representations for locally manipulated images and then integrating them for localization.
Locally manipulated images are more challenging to detect
than fully manipulated images
because the features may not capture the fine details as well.
Our results show that indeed the use of CLIP features to expose locally manipulated images as fakes fails to a large extent.
But we are able to mitigate this problem by equipping the model with a more powerful decoder
that can make better use of the local content.

To validate our method we use the Dolos dataset \cite{tantaru2024}.
This dataset consists of face images whose face attributes (such as mouth, hair, eyes) have been inpainted using 
four methods: two diffusion and two GAN methods.
Due to the small local changes, the narrow domain 
and modern generation methods, many of the images have good perceptual quality.
This is different and more challenging from the images on which CLIP features have generally been applied.
Many of these datasets (such as the one generated with ProGAN \cite{karras2017progressive}, which is used for training) exhibit clear visible semantic artifacts, which may aid generalization.

An intriguing case in Dolos is its subset inpainted with a latent diffusion model (LDM).
The original paper achieved poor results even in the in-domain setting (training and testing on LDM).
The authors speculated that this happens because LDM carries the inpainting in the latent space and the final upscaling step leaves artifacts throughout the image.
We validate this claim by conducting studies on images with clean background.
More importantly, we show that our CLIP-based approach is able to perform localization on the original LDM-inpainted images.
Even more, we observe that training on LDM generalizes well to other generators,
a behaviour that cannot be achieved by training on a different generator or using conventional data augmentation.

Our work makes the following contributions:
(i) We demonstrate that large pretrained representations can effectively be used for deepfake detection and improve generalization over existing methods.
(ii) We present a comprehensive study of the factors that contribute to our model: backbone type, layer, decoder type, and decoder size.
(iii) We achieve high-accuracy manipulation localization in the challenging case of LDM-inpainted images.
Furthermore, we show that training on this type of data improves generalization over other types of data. Our code is available at: \url{https://github.com/bit-ml/DeCLIP}.

\section{Related Work}
\label{sec:related}

In response to advances in generative modelling,
a growing body of research is devoted to exposing fake content;
see \cite{verdoliva2020media,mirsky2021creation,Malik2022DeepFakeDF,nguyen2022cviu, tariang2024synthetic,lin2024detecting} for reviews.
We survey two directions related to our approach,
namely the emerging trend of relying on self-supervised representations for deepfake detection and techniques for the task of deepfake localization.


\mypar{Self-supervised representations in deepfake detection.} Learning transferable representations from unlabelled data has seen impressive progress in recent years \cite{clip,jia21align,dai23instructblip,oquab23dinov2}.
Many of these representations have also been successfully applied to the task of deepfake image detection:
in particular, the CLIP representations \cite{clip} have been the most widely used \cite{Ojha2023TowardsUF,srivatsan23iccv,zhu23gendet,Cozzolino_2024_CVPR,khan24icmr,koutlis2024leveraging,liu24cvpr,reiss24},
but features from other vision-language models (such as BLIP2 \cite{li23blip2} or InstructBLIP \cite{dai23instructblip}) or vision-only models (such as DINO v2 \cite{oquab23dinov2} or MoCo v3 \cite{chen21moco}) have also been employed for deepfake detection \cite{reiss24,chang23antifake,nguyen2024exploring,keit24bilora,you24arxiv}.
These representations are either kept frozen \cite{reiss24} and probed linearly \cite{Ojha2023TowardsUF}, or adapted to the task by full \cite{srivatsan23iccv,khan24icmr} or partial \cite{nguyen2024exploring} fine-tuning, prompt tuning \cite{chang23antifake,khan24icmr}, adapter techniques \cite{khan24icmr,liu24cvpr,keit24bilora}.
The adaptation process can be done most simply by optimising the binary cross-entropy loss \cite{Ojha2023TowardsUF,srivatsan23iccv},
but more recent methods have experimented with the contrastive loss \cite{koutlis2024leveraging},
a teacher-student paradigm \cite{zhu23gendet},
or ways of incorporating the text encoder in the learning process \cite{srivatsan23iccv,khan24icmr,liu24cvpr}.
A similar trend of relying on self-supervised representations
can be noticed for deepfake detection on other modalities:
video \cite{haliassos22realforensics,feng23cvpr,oorloff24cvpr} and
audio \cite{wang22odyssey,oneata_is24,pianese2024training}. 

\mypar{Manipulation localization.}
Local manipulations are the result of
low-level image editing techniques (splicing \cite{dong2013casia} copy-move \cite{wen2016coverage}, object removal \cite{sagong2022rord}) or
deep learning approaches (face swapping \cite{rossler2019faceforensics}, in-painting \cite{yu2019free, repaint}).
Many of the detection approaches rely on a combination of
frequency information \cite{liu2022pscc, wang22cvpr, guo2023hierarchical}
noise information \cite{li2021noise,kwon2022learning,trufor,comprint, zhu24aaai} and
consistency checks \cite{agrawal2022cvprw,huh2018fighting}.
In terms of architecture,
convolutional \cite{wu2019mantranet, liu2022pscc, li2021noise, comprint} and
self-attention \cite{Stehouwer2019OnTD,hao2021transforensics,Das_2022_CVPR,guo2023hierarchical, wang22cvpr} layers are typically employed.
The most common loss is the pixel-level binary cross entropy \cite{hu2020span, kwon2022learning, comprint} or variations such as focal \cite{li2021noise} or Dice loss \cite{trufor};
this is sometimes coupled with an image-level loss \cite{wang22cvpr,yu2024cvprdiffforensics}
or used in a multitask setting \cite{guo2023hierarchical}.
While supervised learning is the typical setup, localisation maps can also be extracted in a weakly-supervised way \cite{tantaru2024},
providing explanations that can help either humans \cite{fosco2022caricatures} or algorithms \cite{Boyd_2023_WACV} improve.
Generalisation is explicitly considered in a few works \cite{kwon2022learning, trufor},
but these focus on the more traditional manipulations of copy-move and splicing.

\section{Overview and preliminaries}
\label{sec:method}


Our goal is to perform localization of manipulated areas in images.
Our approach is based on CLIP features (Sect.~\ref{subsec:clip-features}), since these were shown to yield strong generalization performance for the related task of deepfake detection (classifying if an entire image is fake or real).
However, CLIP features were never evaluated in the context of \emph{locally}-manipulated images. 
Here, we consider the Dolos dataset (Sect.~\ref{subsec:dolos-dataset}), a challenging and carefully-constructed dataset which disentangles multiple axes of image generation.
We perform for the first time (image-level) deepfake detection with CLIP on Dolos (Sect.~\ref{subsec:clip-on-dolos}) and show that CLIP in its original instantiation struggles detecting local images;
we address this problem in the next section.

\begin{table}
 \small
\newcommand\ii[1]{\color{gray} \footnotesize #1}
\centering
\begin{tabular}{rllrr}
\toprule
        &                 &                 & \multicolumn{2}{c}{Test data: Dolos} \\
        & Method          & Train data      & \multicolumn{1}{c}{P2 full}             & \multicolumn{1}{c}{P2 local} \\
\midrule
\ii{1}  & CLIP + linear   & ProGAN          & 93.4                                    & 72.8                         \\
\ii{2}  & CLIP + linear   & Dolos: P2 full  & 98.9                                    & 79.2                         \\
\ii{3}  & Patch Forensics & Dolos: P2 full  & 100.0                                   & 95.3                         \\
\ii{4}  & CLIP + linear   & Dolos: P2 local &  97.2                                   & 71.4                         \\
\bottomrule
\end{tabular}
\caption{%
The impact of full versus local manipulations for detection.
We report the average precision for image-level deepfake detection on the P2 subset from the Dolos dataset.
While CLIP + linear obtains good performance on the fully-generated images from Dolos, it fails to work on locally-generated images.
}
\label{tab:detection-toplines}
\end{table}

\subsection{CLIP features}
\label{subsec:clip-features}

CLIP (contrastive language--image pretraining) \cite{clip} is a foundation vision-language model trained on over 400M image--text pairs scrapped automatically from the web.
Its architecture is composed of two encoders---an image and a text encoder---which are trained to minimize the contrastive InfoNCE loss.
Radford \etal have shown that this model learns visual features that are highly transferable across various tasks.
Recently, Ojha \etal \cite{Ojha2023TowardsUF} have extended this observation by showing that the features extracted from the frozen CLIP image encoder can discriminate between fake and real images.
The simple approach of applying a linear classifier on CLIP features works well not only in-domain,
but, more importantly, it generalizes better than prior work on a number of different datasets,
such as diffusion-generated image, video data, low-level image manipulations.
Among the image encoder architectures provided by CLIP,
Ojha \etal have shown that the visual transformer \cite{vit} performs better than the residual network \cite{he2016deep}.

\subsection{Dolos dataset}
\label{subsec:dolos-dataset}

Dolos \cite{tantaru2024} is a recently introduced dataset of locally manipulated faces.
The dataset has been used to analyse the capabilities of weakly-supervised deepfake methods,
and as such it provides a controlled setup over three components of image generation:
inpainting type (local, full),
model family (P2~\cite{P2}, LDM~\cite{rombach2022ldm}, LaMa~\cite{LAMA}, Pluralistic~\cite{pluralistic}),
and training data (CelebA-HQ, FFHQ).
We use the inpainting type information to study the effect of local manipulations (Sect.~\ref{subsec:clip-on-dolos}) and
the model family information to study generalization across generators (Sect.~\ref{subsec:main-results}).
Regarding the generator training data, we restrict ourselves to the CelebA-HQ variants.
The generated images (especially those produced by diffusion models---P2 and LDM) are highly realistic,
making Dolos a challenging out-of-domain dataset.

\subsection{Detection with CLIP on Dolos}
\label{subsec:clip-on-dolos}


The majority of datasets considered by Ojha \etal \cite{Ojha2023TowardsUF} are fully-generated images.
But how does the CLIP-based model of \cite{Ojha2023TowardsUF} perform on partially-manipulated images?
To answer this question, we consider images from Dolos inpainted with the P2 diffusion model,
for which we have both fully and locally generated images.
We report average precision for image-level detection.
Table \ref{tab:detection-toplines} shows the results for multiple combinations of methods and training data.

First, we observe that applying the original method of Ojha \etal (CLIP + linear trained on ProGAN) on fully-generated images from Dolos yields an average precision of 93.4\% (row: 1, col: P2 full).
This performance is similar to the average performance of 93.3\% reported in Table 2 from \cite{Ojha2023TowardsUF},
which indicates good generalization, as we move to a different domain (from general images to faces) and to a different generator (from GAN to diffusion).

However, the pretrained CLIP + linear model does not work as well on local manipulations,
as the performance drops from 93.4\% to 72.8\% (row 1).
This conclusion is also supported by the results on face swap manipulations (another type of local manipulations):
Ojha \etal report 82.5\% AP (Table 9, col: 9, ``Deepfakes''), which is the second lowest performance out of the 19 datasets used therein.

Importantly, the performance on local manipulations is not improved even if we train on in-domain data:
training on either P2 full or P2 local still yields only 79.2\% (row 2) or 71.4\% (row 4), respectively.
On the other hand, Patch Forensics \cite{patch-forensics}, which was used as a baseline in \cite{tantaru2024}, is not affected local manipulations:
it achieves an average precision of 95.3\% (row 3).
This result serves as our motivation for developing a patch-based approach on CLIP features.
The full model, which we describe in the next section, is able to match the performance of Patch Forensics on local images, while maintaining good generalization performance.


 
\section{Deepfake localization with CLIP}
\label{sec:exp}


Given an image that has been manipulated locally, our aim is to produce a map showcasing where the manipulation has occurred: values close to 1 indicate that the corresponding pixel has been altered; values close to 0 indicate that the pixel is authentic. We assume a fully supervised setting in which we have access to images and groundtruths maps; this setup is reminiscent to the one encountered in the object segmentation task. 

Our main idea is to leverage high quality pretrained image representations and couple them with an appropriate decoder trained for deepfake manipulation localization.
This is achieved by two components:
an image encoder, which encodes the image as a low-resolution grid of features,
and a decoder, which upscales the encoded representations to the higher-resolution of the input image.
The resulting method, which we name \ours, is shown in Figure \ref{fig:overview}.


\textbf{Encoder.}
We extract representations from both pretrained CLIP image architectures (visual transformer and residual network) at various layers.
For the visual transformer, we choose the ViT-L/14 variant,
which operates on 16$\times$16 patches of size 14$\times$14 and has 24 self-attention layers;
each layer outputs 256 1024-dimensional embeddings of the input patches and one additional global CLS token, which we discard. 
For the residual network, we use the ResNet-50 variant. 
This variant has four blocks:
after the first block the output is a 56$\times$56 256-dimensional embedding;
with each subsequent block the embedding dimension doubles, while the spatial resolution halves.


\textbf{Decoder.}
We decode the information from the CLIP representations using a convolutional-based architecture.
This architecture consists of four blocks,
each sequencing $M$ sub-blocks and a $\times 2$ bilinear upsampling layer.
A sub-block is composed of a 5$\times$5 convolutional layer followed by batch normalization and ReLU activations.
To project the output to the grayscale mask space,
we use a final 5$\times$5 convolutional layer.
The decoders mentioned throughout the paper are conv-\{4, 12, 20\}, where the number indicates the total number of sub-blocks ($4M$) and controls the decoder size.

The choices regarding the encoder backbone, layer at which features are extracted, decoder size are analysed in Sects.~\ref{subsec:main-results} and \ref{subsec:ablations}.


\subsection{Experimental setup}

\label{subsec:experimental-setup}

\begin{table}
    \newcommand\ii[1]{\color{gray} \footnotesize #1}
    \centering
    \small
    \begin{tabular}{rllrr}
        \toprule
        & \multicolumn{2}{c}{Method}            & \multicolumn{2}{c}{IoU $\uparrow$} \\
        \cmidrule(lr){2-3}
        \cmidrule(lr){4-5}
        & Backbone & Decoder      & ID         & OOD        \\
        \midrule
        & \multicolumn{4}{l}{\textit{Patch Forensics variants} \cite{patch-forensics, tantaru2024}} \\
        \ii{1} & Xception (L2) & linear & 69.3 & 20.4 \\
        \ii{2} & Xception (L2) & conv-20 & 70.4 & 12.6 \\
        \midrule
        & \multicolumn{4}{l}{\textit{CLIP variants} \cite{Ojha2023TowardsUF}} \\
        \ii{3} & ViT-L/14 (L24) & linear & 36.2 & 18.5 \\
        \midrule
        & \multicolumn{4}{l}{\it Other methods} \\
        \ii{4} & \multicolumn{2}{l}{PSCC-Net \cite{liu2022pscc}}         & \bf 81.0 & 21.6  \\
        \ii{5} & \multicolumn{2}{l}{CAT-Net \cite{kwon2022learning}} & 18.5     & 17.9  \\
        \midrule
        & \multicolumn{4}{l}{\it \ours variants} \\
        \ii{6} & ViT-L/14 (L21)               & conv-20 & 67.9 & 32.6 \\
        \ii{7} & RN50 (L3)                    & conv-20 & 71.0 & 32.0 \\
        \ii{8} & ViT-L/14 (L21) $+$ RN50 (L3) & conv-20 & 73.8 & \textbf{34.7} \\
        \bottomrule
    \end{tabular}
    \caption{%
    Comparison of deepfake localization methods on Dolos.
    We report in-domain (ID) IoU (averaged over the four datasets in Dolos)
    and out-of-domain (OOD) IoU (averaged over the twelve train--test combinations, where train and test sets differ). 
    \ours performs better in the OOD scenario, while also showing good ID performance, being outperformed only by PSCC-Net. 
    }
    \label{tab:main}
\end{table}

\begin{figure*}[htb!]
\includegraphics[trim={0 0.2cm 0 0.2cm},clip,width=\textwidth]{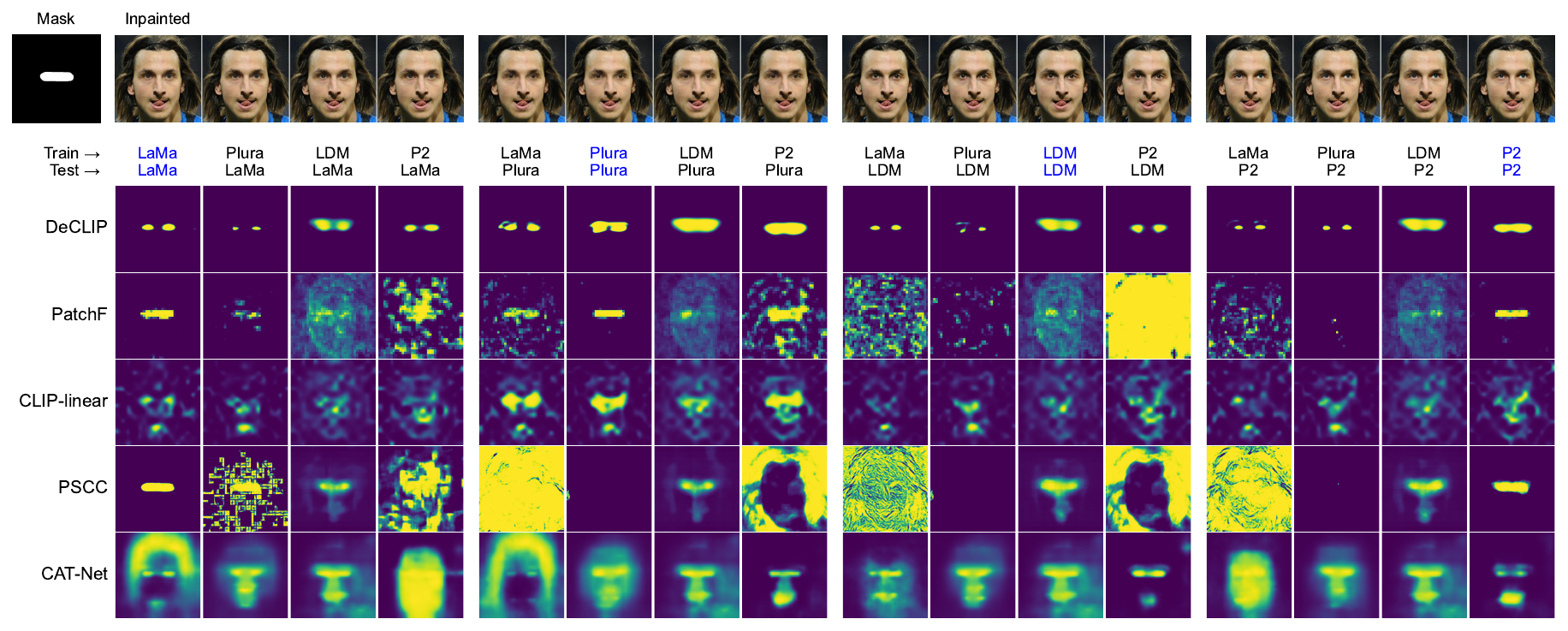}
\caption{%
Sample predictions for \ours (second row) and four other methods (Patch Forensics, CLIP-linear, PSCC, CAT-Net) on all 16 train--test combinations from the Dolos dataset.
The in-domain combinations are highlighted in blue; the others are out-of-domain combinations.
The black-and-white image in the top left corner shows the inpainting mask (white is the inpainted region) and the rest of the images in the first row are the inpainted images with one of the four test datasets (LaMa, Pluralistic, LDM, P2).
}
\label{fig:qualitative}
\end{figure*}

\mypar{Dataset and metrics}.
We report results on the four locally-manipulated subsets from the Dolos dataset (Sect.~\ref{subsec:dolos-dataset}): LaMa, Pluralistic, LDM, P2. We consider all 16 train-test combinations (4 train $\times$ 4 test)
and report the intersection over union (IoU) between the predicted binary mask and the groundtruth mask.
We obtain binary predictions by using a fixed threshold of 0.5 over the continuous predictions.
To ease comparison between methods, we report aggregated metrics.
We consider the averaged IoU based on whether the train and test dataset match.
\textbf{ID IoU} (in-domain intersection-over-union) is computed as the average IoU over the 4 combinations when the training and test sets match.
It serves as a topline, measuring the difficulty of the chosen datasets
(how well the detector can learn patterns inflicted by a fixed deepfake generator).
\textbf{OOD IoU}  (out-of-domain intersection-over-union) is computed as the average IoU over the 12 combinations when the training and test sets differ.
It measures the generalization to unseen data and the model's capability to handle diverse variations in the data.

\mypar{Implementation details.}
We adapt the training setup for deepfake detection from \cite{Ojha2023TowardsUF} for localization.
We optimize the binary cross-entropy loss between the predicted and groundtruth masks.
The hyper-parameters are kept the same.
Specifically, we use the Adam optimizer with an initial learning rate of $10^{-3}$,
which is reduced by a factor of 10 using a patience of five epochs.
The training is stopped when the learning rate decreases under $10^{-6}$.


\subsection{Main results}
\label{subsec:main-results}

\mypar{Baselines.}
We compare \ours against the following baselines:
(i) Patch Forensics was proposed in \cite{patch-forensics} and used in \cite{tantaru2024} for weakly-supervised localization and cross-generator localization on the Dolos dataset.
The original method extracts features from block 2 of the Xception network and
projects them to binary predictions using a 1$\times$1 convolutional layer.
Since this last step is equivalent to a linear decoder,
we also experiment with a stronger 20-layer convolutional decoder.
(ii) CLIP:ViT-L/14-linear is the method proposed in \cite{Ojha2023TowardsUF} for image-level detection, which we adapt minimally for localization:
instead of using the CLS token we use the feature map extracted at the last (L24) layer of ViT-L/14 encoder and learn a linear patch classifier on top.
(iii) PSCC-Net \cite{liu2022pscc} learns to extract local and global features from images and estimates manipulation masks at multiple scales.
(iv) CAT-Net \cite{kwon2022learning} uses discrete cosine transform coefficients to learn compression artifacts to localize image manipulations.

\begin{figure}
\centering
\includegraphics[trim={0 1cm 0 0.9cm},clip,width=0.45\textwidth]{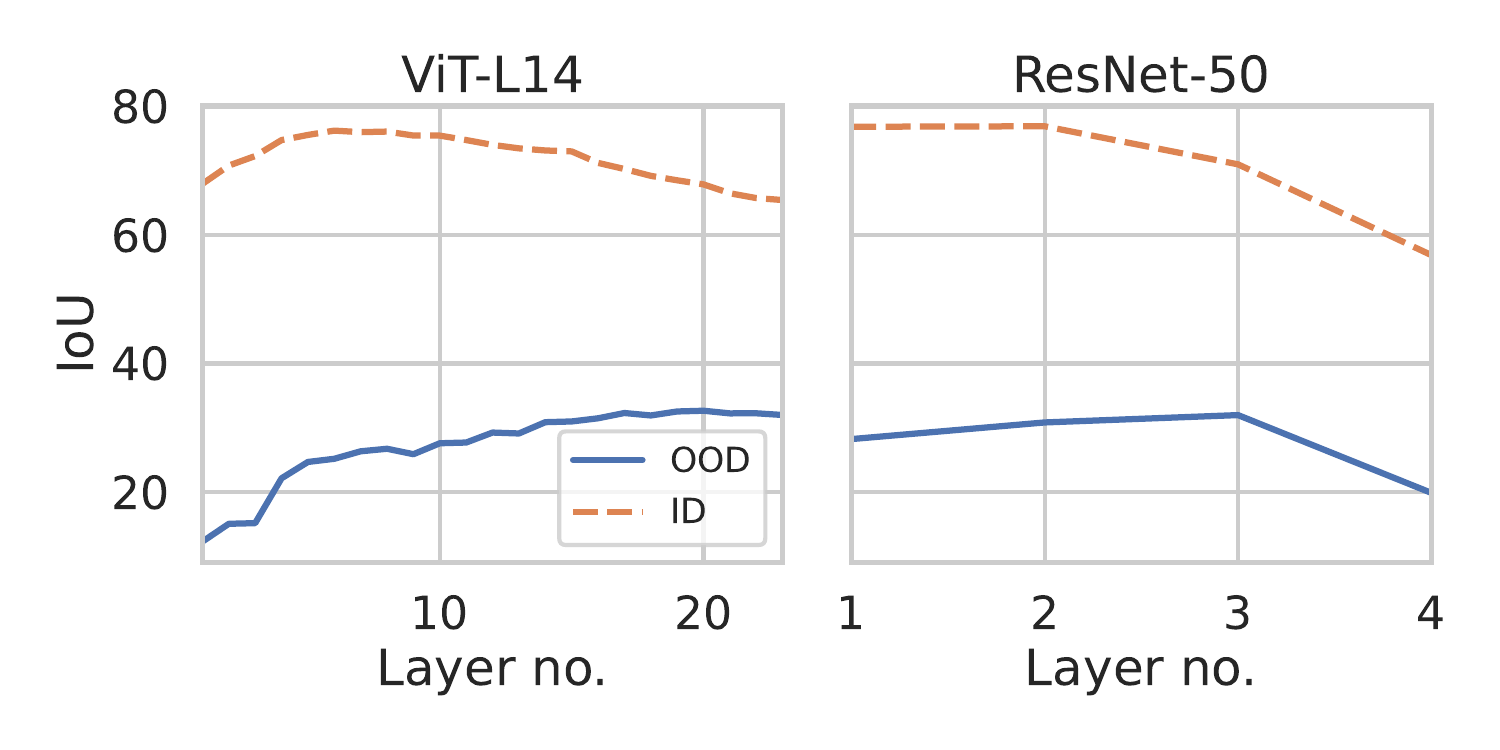}
\caption{%
    The impact of the layer at which the features are extracted for the ViT-L/14 (left) and ResNet-50 (right) backbone.
    We report IoU performance on the Dolos dataset both in-domain (ID, orange dashed line) and out-of-domain (OOD, blue solid line).
    }
\label{fig:layers}
\end{figure}

\mypar{Quantitative results.}
Our main results are shown in Table \ref{tab:main}. For all methods we use all 16 train-test combinations of generators in Dolos and average IoU results for ID and OOD setups.
Patch Forensics (row 1), the method that was originally applied for this task, shows good ID performance but performs poorly in OOD.
Our starting point, the original CLIP method (row 3) has much worse performance in ID than Patch Forensics and comparable performance in OOD. Rows 6--8 show variants of our method, \ours, which bring a significant boost to the original CLIP, improving both in ID and OOD setups. Compared to Patch Forensics, it has similar good performance in ID, but a 50\% relative improvement in OOD.
We also experiment with adding a larger decoder to Patch Forensics (row 2), but it did not help improve the OOD performance.
Rows 4 and 5 show comparisons with other methods, PSCC-Net and CAT-Net that have been re-trained and tested in the exact same scenarios as \ours.
Both have significantly lower performance in OOD compared to \ours.
Their behaviour is different in ID: PSCC has good performance, while CAT-Net very poor. 


\mypar{Qualitative results.} We show examples of output localization masks for all train--test setups produced by \ours (ViT-L/14) and other methods in Figure \ref{fig:qualitative}.
Notice that \ours produces more consistent and clean masks even in the harder, OOD scenarios.
PSCC-Net and Patch Forensics generally perform well in ID (with the exception of LDM--LDM case), but struggle in the OOD scenarios.
CAT-Net  and CLIP:ViT-L/14-linear seem to learn general face features,
not related to the actual inpainted region.

\subsection{Ablations}
\label{subsec:ablations}

\mypar{Backbones and representations depth.} 
For both ViT-L/14 and ResNet-50 backbones we vary the depth of the layer at which we extract the pretrained representations.
Our results are shown in Figure \ref{fig:layers}.
When using ResNet-50, representations extracted at lower convolutional blocks (L1, L2) are best for manipulation localization in ID, while representations extracted at L3 block is best for OOD.
The last block, L4 has lower performance both in ID and OOD.
In the case of ViT-L/14, a similar trend can be seen with ID localization being higher when using features extracted at lower layers (L7), while OOD localization accuracy increasing when using higher level features (L21).
Unlike  ResNet-50, for ViT-L/14 there is no significant drop in performance at the last layer in the OOD scenario. 

%

\begin{figure}
    \centering
    \includegraphics[trim={0 0.3cm 0 0.0cm},clip,width=0.48\textwidth]{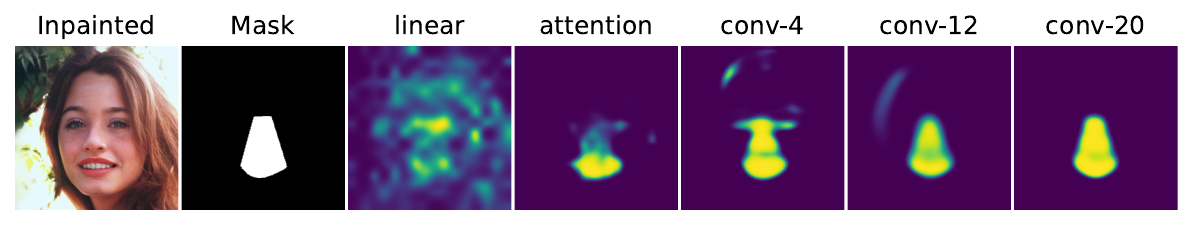} \\
    \includegraphics[trim={0 0.2cm 0 0.3cm},clip,width=0.48\textwidth]{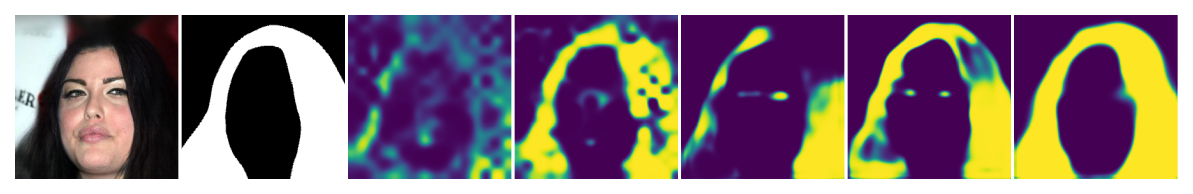}
    \caption{
    Predicted masks obtained with different decoders.
    All results use \ours ViT-L/14 variant.
    First row shows the LDM--P2 scenario, while the second P2--LaMa.
    The larger convolutional decoder produces more smooth and precise results.}
    \label{fig:decoders}
\end{figure}

\mypar{Decoder architecture.}
We experiment with three types of decoders: linear, convolutional and self-attention.
For the convolutional one we vary the depth and choose 4, 12 and 20 sub-blocks.
The self-attention decoder has 2 attention blocks. Each attention block has 16 heads, a hidden size of 1024, associated with a MLP of size 4096.
For all decoders, we use bilinear upsampling. 
Results are shown in Table \ref{tab:decoders} for \ours with ViT-L/14 backbone.
The convolutional decoder outperforms both the linear and the self-attention one.
Moreover, the larger the decoder, the better the performance both in ID and OOD scenarios.
This indicates that localization manipulation needs larger decoders to properly make use of pretrained representations.
Visual examples are shown in Figure \ref{fig:decoders} for two train-test scenarios: LDM--Pluralistic and P2--LaMa.
The identified manipulation mask becomes more precise (less erosion, fewer holes) as we move from the linear decoder to the attention-based one and convolutional decoders.  

\begin{table}[htb!]
\centering
  \footnotesize
 \begin{tabular}{llrrr}
        \toprule
         \multicolumn{2}{c}{Method}      &       & \multicolumn{2}{c}{IoU $\uparrow$} \\
        \cmidrule(lr){1-2}
        \cmidrule(lr){4-5}
         Backbone & Decoder  & Params.  & ID         & OOD        \\
        \midrule
 ViT-L/14 & linear    & $1.0 \times 10^3$  & 39.9 & 19.2\\
 ViT-L/14 & attention & $25.1 \times 10^6$ & 61.8 & 27.9 \\
 ViT-L/14 & conv-4    & $17.4 \times 10^6$ & 65.3 & 28.6 \\
 ViT-L/14 & conv-12   & $34.8 \times 10^6$ & 66.8 & 31.0 \\
 ViT-L/14 & conv-20   & $52.2 \times 10^6$ & \textbf{67.9} & \textbf{32.6} \\
 \bottomrule
\end{tabular}
\caption{%
    Influence of decoder type and size on manipulation localization with \ours ViT-L/14.
    The convolutional decoder outperforms linear and self-attention ones.
    The larger decoder (with 20 convolutional sub-blocks) performs best in both ID and OOD.
}
\label{tab:decoders}
\end{table}



\subsection{Detailed results}
\label{subsec:detailed-results}

\begin{figure}
    \includegraphics[trim={0 0.2cm 0 0.2cm},clip,width=\linewidth]{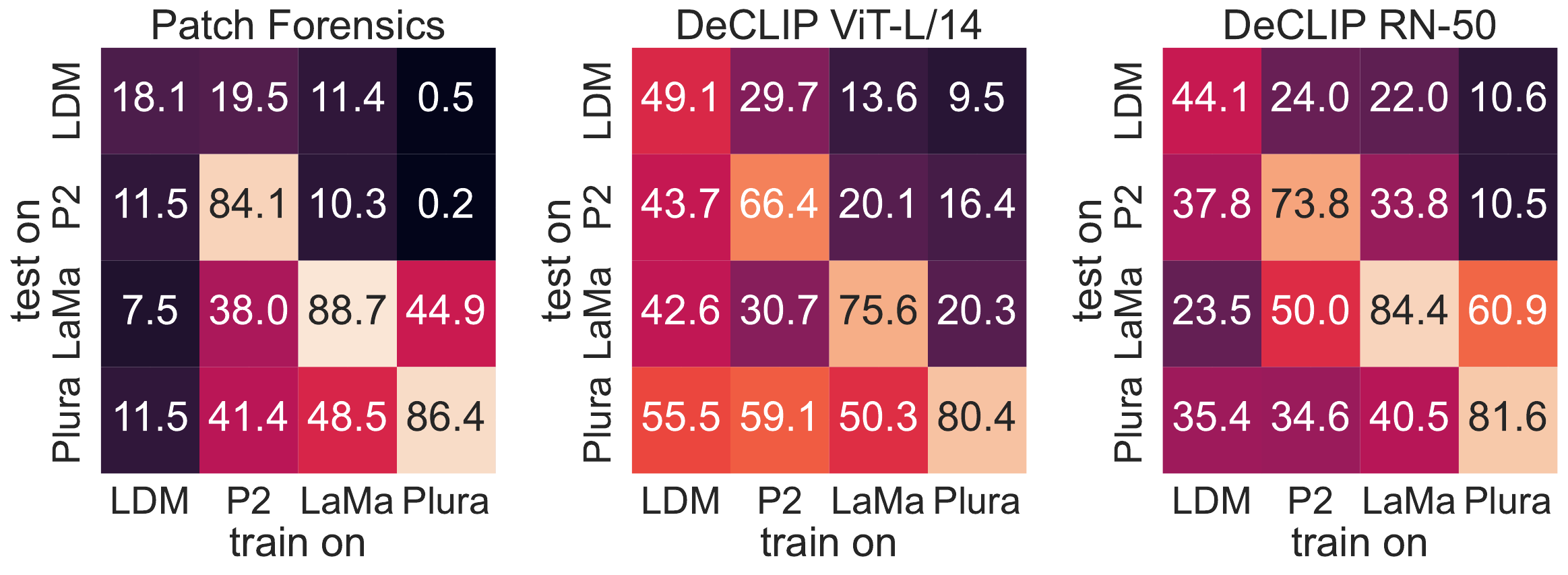}
    \caption{%
    Detailed cross-generator performance on the Dolos dataset for three methods:
    Patch Forensincs \cite{tantaru2024},
    \ours with ViT-L/14 backbone at layer 21,
    \ours with ResNet-50 backbone at layer 3.
    Both \ours variants use the conv-20 decoder.
    }
    \label{fig:detailed_results}
\end{figure}

In the previous sections, in order to summarize the 
localization performance of different models we used 
aggregate measures over in-domain and out-of-domain train--test combinations.
Here we provide a more detailed view by showing the results of each train--test combination.
Figure \ref{fig:detailed_results} 
shows these results for the Patch Forensics method 
\cite{tantaru2024} 
and for \ours with either ViT-L/14 or ResNet-50 backbone. The diagonal of these cross-generator matrices shows the in-domain performance for each dataset.
We see that Patch Forensics is slightly more accurate for three of the generators (P2, LaMa, Pluralistic), but it fails completely on the LDM generator.
\ours, on the other hand, is more stable on all four datasets and even gives good results on LDM data (44.1\% and 49.1\%, respectively). Looking at the columns, we can see how well one dataset transfers to the others.
Interestingly, we see that when training on LDM,  \ours also generalizes well to other test datasets.
This is not the case when training on Pluralistic, whose performance on P2 and LDM is low for all the methods shown; this suggests that Pluralistic fingerprints have very little in common with those produced by diffusion-based generators (P2 or LDM).
The case of LDM is worth further investigation, which we do in the next section (Sect.~\ref{sec:ldm_sec}).
Finally, we observe the complementarity of the two \ours variants.
Even if on average their in-domain and out-of-domain IoUs are similar (see Table \ref{tab:main}, rows 6--7), there are train-test combinations 
where 
the two backbones show contrasting behaviours: for example, from Pluralistic to LaMa, the ResNet backbone performs better (60.9 versus 20.3); from P2 to Pluralistic, ViT performs better (59.1 versus 34.6).
For this reason, concatenating the representations from the two backbones improves both ID and OOD performance compared to their individual results (see Table \ref{tab:main}, row 8).

\section{The case of LDM-inpainted images} 
\label{sec:ldm_sec}


\begin{table*}
        \newcommand\ii[1]{\color{gray} \footnotesize #1}
        \centering
        \footnotesize
        \begin{tabularx}{\textwidth}{rXccccrrrrrr}
                \toprule
                &
                & \multicolumn{2}{c}{Out region}
                & \multicolumn{2}{c}{In region}
                & \multicolumn{3}{c}{Test data: LDM variants}
                & \multicolumn{3}{c}{Test data: OOD datasets}
                \\
                \cmidrule(lr){3-4} \cmidrule(lr){5-6} \cmidrule(lr){7-9} \cmidrule(lr){10-12}
                          & Train     & Content & Fingerprint & Content & Fingerprint & LDM/real & LDM/clean & LDM      & P2       & LaMa     & Plura \\
                \midrule                                                                                    
                \ii{1}    & LDM/real  & real    &             & real    & \checkmark  & \bf 62.1 & 62.2      & 24.5     & 56.9     & 23.2     & 43.2 \\
                \ii{2}    & LDM/clean & real    &             & fake    & \checkmark  & 39.7     & \bf 67.3  & 38.3     & \bf 59.6 & 27.4     & \bf 56.8 \\
                \ii{3}    & LDM       & real    & \checkmark  & fake    & \checkmark  & 17.1     & 53.1      & \bf 49.1 & 43.7     & \bf 42.6 & 55.5 \\
                \bottomrule
        \end{tabularx}
        \caption{%
                The impact of the LDM fingerprint on the inpainted (in) and background (out) regions.
                We report IoU using \ours with ViT-L/14 backbone, layer 21 and conv-20 decoder.
        }
        \label{tab:ldm-fingerprint-analysis}
\end{table*}

We have seen in Sect.~\ref{subsec:detailed-results} that localizing  manipulations in images inpainted with LDM is more challenging than performing this task on images inpainted with other techniques (P2, LaMa, Pluralistic).
Furthermore, we observed that training \ours on LDM data gives a strong out-of-domain performance.
What is the reason for this?


First, we recall that LDM provides an atypical case of image inpainting.
Unlike the other three inpainting methods considered,
LDM inpainting takes place in the \emph{latent} space.
As such, the generated latent image must be projected back to the pixel space.
This upscaling step is performed by a variational autoencoder (VAE) network,
which leaves artifacts throughout the entire generated image,
not just in the inpainted regions as for the other three methods.
These artifacts, although imperceptible, are detectable by the networks and is what makes localization challenging.
In what follows we conduct multiple analyses to understand:
(i) the impact of the model capacity;
(ii) the impact of the fingerprint left by LDM on the background;
(iii) the relationship between the LDM fingerprint and data augmentation;
(iv) the performance on general content images.

\begin{table}
\centering
\footnotesize
\tabcolsep 5pt
\begin{tabular}{lrrrrr}
\toprule
           & Params. & \multicolumn{4}{c}{Test data} \\
           \cmidrule(lr){3-6}
Method     & $\times 10^6$ & LDM & P2 & LaMa & Plura \\
\midrule
PatchF. (linear) \cite{patch-forensics} &  0.2 &     18.1 &     19.5 &     11.4 &      0.5 \\
PatchF. (conv-20)                       & 42.6 &     39.2 &     25.6 &     18.5 &     24.1 \\
PSCC \cite{liu2022pscc}                 &  3.6 &     41.5 &     23.6 &     26.2 &     27.6 \\
\ours                                   & 52.2 & \bf 49.1 & \bf 43.7 & \bf 42.6 & \bf 55.5 \\
\bottomrule
\end{tabular}
\caption{%
    Comparison in terms of IoU of localization methods trained on LDM.
    \ours uses features from ViT-L/14 layer 21 and conv-20 decoder.
}
\label{tab:ldm-methods}
\end{table}


\mypar{Larger models improve performance on LDM, but capacity alone is not sufficient for generalization.}
Patch Forensics fails at localizing even ID manipulations on LDM images, while \ours works better.
An important difference between the two is the larger decoder employed by the latter.
We verify whether the reason for the difference in performance is solely based on network capacity.
We consider two larger variants:
 Patch Forensics but with the conv-20 decoder (42.6M parameters) and  PSCC \cite{liu2022pscc} (3.6M parameters);
both of these networks are trained from scratch on the LDM subset.
The results in Table \ref{tab:ldm-methods} show that indeed the network capacity is responsible to a degree for the good performance, as the larger variant of Patch Forensics improves substantially over the smaller baseline.
However, PSCC is better than both Patch Forensics variants,
while \ours achieves the best performance, at a number of parameters comparable to the larger Patch Forensics variant.

\mypar{LDM background fingerprint provides stable out-of-domain performance.}
In-domain performance on LDM is lower than that on the other three generation methods.
Conversely, the generalization performance of LDM is much stronger than that of the other generation methods.
We investigate the role played by the LDM fingerprint in this behaviour.
To disentangle this aspect, we create two variants of LDM datasets:
\begin{itemize}
    \item LDM/clean, which uses a fingerprint-free background.
    This variant is created by replacing the background (the complement of the mask) of the LDM-generated images with information from the original real image.
    \item LDM/real, which consists of real images with fingerprint on the masked region.
    This variant is created by passing the real images through LDM using an empty mask and then cleaning the background.
\end{itemize}

The results are shown in Table~\ref{tab:ldm-fingerprint-analysis}.
Cleaning the background fingerprint improves in-domain results:
from 49.1 to 62.1 and 67.3 (see the diagonal along the ``LDM variants'' columns);
this is to be expected since there are no distractors on the background.
Relying solely on the fingerprint information (row 1) gives good results on two of the out-of-domain datasets (56.9 on P2 and 43.2 on Plura),
suggesting that LDM shares a similar fingerprint to these methods.
Conversely, the poor result on LaMa (23.2) indicates that its fingerprint is different.
By further manipulating the content of the target region (row 2), we notice stronger results on all three datasets.
A possible reason is that this setup is similar to that of the other datasets:
the background is clean, while the manipulated region is affected by both low-level and semantic changes.
However, the original LDM (row 3) ensures the most consistent out-of-domain performance, improving the performance on LaMa---the most challenging dataset---from 27.4 to 42.6.
This might happen because LDM forces the model to disregard the fingerprint and focus on semantic information, which is more transferable.

\mypar{Low-level data augmentations induce a similar, but weaker effect than the LDM fingerprint.}
A possible explanation for the improved generalization showcased when training on LDM-inpainted data is that the VAE decoding acts as data augmentation:
it introduces low-level artifacts on the entire images,
forcing the detection model to be robust to low-level changes and more aware to semantic inconsistencies.
This observation raises the question:
Would a different low-level data augmentation help with generalization?
We experiment with three types of augmentations (Gaussian blur, color jitter and JPEG compression), which we apply on all images from the LDM/clean dataset.
The results in Table~\ref{tab:ldm-clean-data-aug} show the augmentations have a similar effect to that produced by the LDM fingerprint:
they level up the results across datasets, by improving performance on the LDM and LaMa datasets, while sometimes hurting performance on P2 and Pluralistic.
However, none of the augmentations nor their combination helps on average as much as the fingerprint artifacts present in LDM.


\begin{table}
 \footnotesize
\centering
\newcommand\ii[1]{\color{gray} \footnotesize #1}
\begin{tabular}{llrrrr}
  \toprule
  &                &         & \multicolumn{3}{c}{AutoSplice} \\
  \cmidrule(lr){4-6}
  & Method         & COCO-SD & 75       & 90       & 100 \\
  \midrule
  & \multicolumn{5}{l}{\textit{Pretrained models}} \\
\ii{1} & PSCC \cite{liu2022pscc}           & \textbf{8.2}     &  1.8     & 3.6      & 48.2 \\
\ii{2} & CAT-Net \cite{kwon2022learning}        & 0.1     & \textbf{33.1}      &\textbf{69.7}      &\textbf{76.1} \\
\ii{3} & TruFor\cite{trufor}  & 6.7	 &  20.7   &    34.1  &   55.6    \\
\ii{4} & MantraNet \cite{wu2019mantranet}   &  2.0		  &   3.2	  &   3.2   &  13.1    \\
  \midrule
  & \multicolumn{5}{l}{\textit{Models trained on COCO-SD}} \\
\ii{5} & Patch Forensics & 16.4    & 28.1  & 28.4     & 28.2 \\
\ii{6} & PSCC           & 33.4    & 48.5   & \bf 50.6   & 49.4 \\
\ii{7} & CAT-Net        & 15.5    &  0.6   &  0.8     &  4.7 \\
\ii{8} & \ours          &\textbf{51.1}     & \bf 49.6     & 49.9     & \textbf{50.1} \\ 
  \bottomrule
\end{tabular}
\caption{%
    Localization performance (in terms of IoU) on the general domain of MS COCO.
    \ours shows stable performance across all datasets.
    Bold indicates best results in each of the sections.
}
\label{tab:ldm-general-domain}
\end{table}


\begin{table}
\centering
 \footnotesize
\tabcolsep 4pt
\begin{tabular}{lcccrrrr}
\toprule
           & \multicolumn{3}{c}{Augmentation} & \multicolumn{4}{c}{Test data} \\
          \cmidrule(lr){2-4} \cmidrule(lr){5-8}
Train data & Blur & Color & JPEG & LDM  & P2 & LaMa & Plura \\
\midrule
LDM/clean  &            &              &            & 38.3 & \bf 59.6 & 27.4 & \textbf{56.8} \\
LDM/clean  & \checkmark &              &            & 43.9 & 52.5 & 33.9 & 55.4 \\
LDM/clean  &            & \checkmark   &            & 39.7 & \textbf{59.6} & 23.0 & 54.4 \\
LDM/clean  &            &              & \checkmark & 47.0 & 53.9 & 35.4 & 43.4 \\
LDM/clean  & \checkmark & \checkmark   & \checkmark & 43.9 & 48.7 & 33.4 & 49.3 \\
\midrule
LDM        &            &              &            & \textbf{49.1} & 43.7 & \textbf{42.6} & 55.5 \\
\bottomrule
\end{tabular}
\caption{%
    Data augmentation on LDM/clean.
    We report IoU on Dolos for the \ours model trained on LDM/clean augmented with either blur, color jitter or JPEG compression.
}
\label{tab:ldm-clean-data-aug}
\end{table}

%

\begin{figure}
    \centering
    \includegraphics[trim={0 0.3cm 0 0.0cm},clip,width=0.48\textwidth]{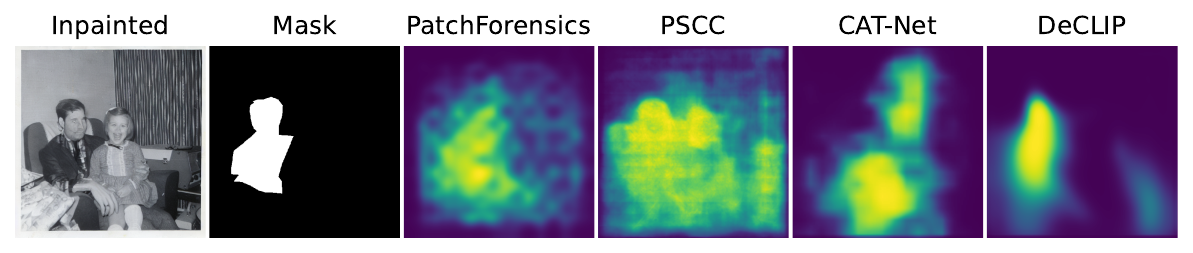} \\
    \includegraphics[trim={0 0.2cm 0 0.3cm},clip,width=0.48\textwidth]{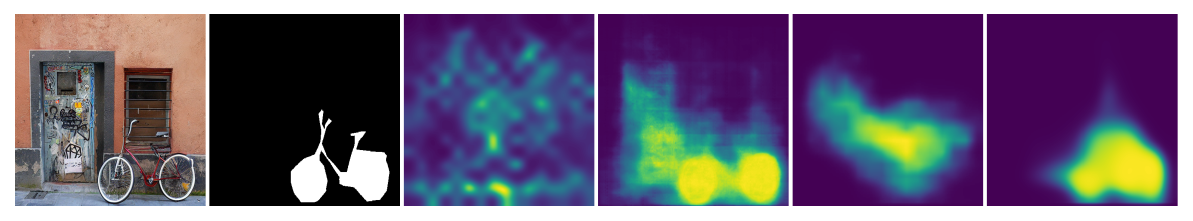}
    \caption{%
    Manipulation localization on COCO-SD.
    \ours offers a more precise localization 
    than the other methods.}
    \label{fig:cocosd}
\end{figure}

\mypar{Conclusions translate to more general domains.}
While the Dolos dataset enabled a careful analysis of deepfake localization in a challenging and realistic setting,
it covers only a narrow domain: faces.
We verify whether the main conclusions apply to more general images.
We inpaint almost 11k images
(9k train, 829 validation, 985 test) from MS COCO \cite{lin2014microsoft} with Stable Diffusion \cite{rombach2022ldm}, which is also an LDM model.
We select the mask of a random object whose area is larger than 5\% and
prompt the inpainting model with the original image caption.
This dataset, which we name COCO-SD, is used for training the localization models. 
(A similar dataset has been proposed by concurrent work \cite{mareen2024tgif}.)
To evaluate generalization, we use the AutoSplice dataset \cite{hia2023autosplice}.
This dataset consists of images manipulated by DALLE-2 \cite{ramesh2022dalle2}, where specific 
objects are replaced with other objects.
It has three variants differing in terms of 
JPEG compression:
the original uncompressed one (quality factor: 100)
and two compressed ones (quality factors of 75 and 90).
Table~\ref{tab:ldm-general-domain} presents the results for \ours, Patch Forensics, PSCC and CAT-Net.
We also show results of pretrained models
for the two last methods, as well as for MantraNet \cite{wu2019mantranet} and TruFor \cite{trufor}.
We observe that the pretrained models, especially CAT-Net and PSCC, are unstable:
they have good performance on some of the datasets, but equally poor on others.
MantraNet performs poorly on all datasets, while TruFor has good performance on AutoSplice-100 but lower on 
its JPEG compressed variants and COCO-SD.
Instead, training on the LDM-based COCO-SD dataset offers more stable performance.
This is especially true for Patch Forensics, PSCC and \ours,
which perform similarly across all three AutoSplice variants.
In terms of model comparison, \ours works better in-domain,
while slightly outperforming PSCC on the out-of-domain AutoSplice dataset.
The visual examples in Figure \ref{fig:cocosd} show that
\ours produces more precise localization of the forged object.

%

\section{Conclusion}
Our paper presented \ours---a first attempt at decoding large self-supervised representations for manipulation localization.
Through extensive experiments we showed that, not only is manipulation localization feasible using these features, but they also significantly improve generalization capabilities in OOD scenarios, when there is a train--test generator mismatch.
We conducted a comprehensive analysis of the factors that contribute to the successful decoding of those features: backbone type, layer depth, decoder type and size.
We found that larger, convolutional decoders improve the quality of the predicted masks compared to linear or self-attention ones.
Moreover, VIT-L/14 and ResNet-50 backbones show contrasting behavior 
that can be exploited by combining representations from both backbones.
Finally, we showed that, contrary to prior assumptions, manipulation localization can be effectively performed even in the challenging case of LDM data.
Interestingly, learning on this type of data offers robustness and improves generalization to other types of local manipulations. 

\mypar{Acknowledgements.}
This work was supported in part by the EU Horizon projects AI4TRUST (No. 101070190) and ELIAS (No. 101120237),
and by CNCS-UEFISCDI (PN-IV-P7-7.1-PTE-2024-0600).

{\small
\bibliographystyle{ieee_fullname}
\bibliography{bibliography}
}

\include{Supplementary_2}

\end{document}


\title{DeCLIP: Decoding CLIP representations for deepfake localization\\ Supplementary material}

\maketitle



\section{Additional detailed results}

\begin{figure}[htb!]
    \includegraphics[trim={0 0.2cm 0 0.2cm},clip,width=\linewidth]{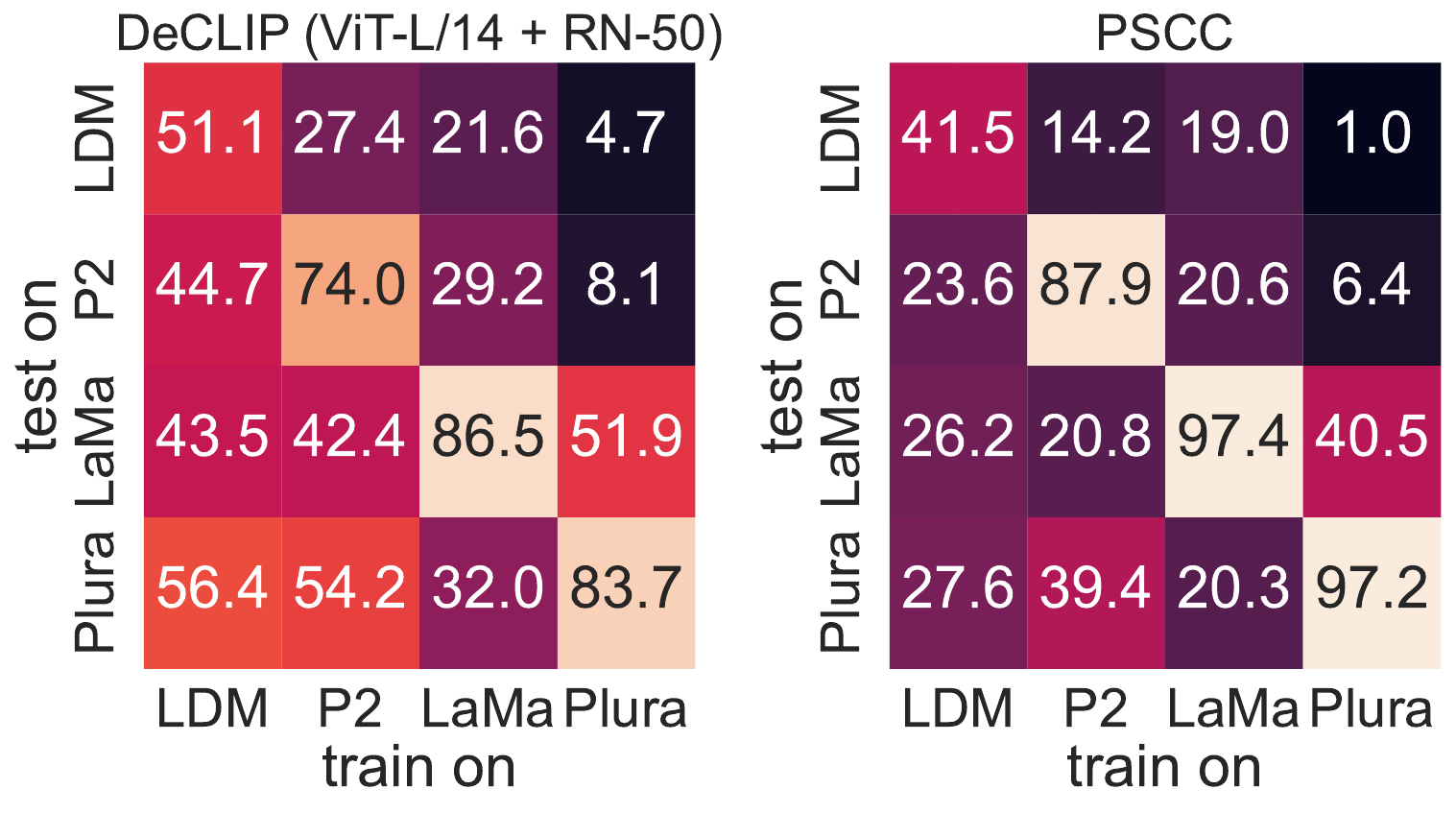}
    \caption{%
    Detailed cross-generator performance (IoU) on the Dolos dataset (all 16 train--test combinations) for \ours that used both ViT-L/14 and ResNet-50 representations and PSCC. 
    }
    \label{fig:detailed_results}
\end{figure}

In Figure \ref{fig:detailed_results} we provide additional detailed results on all train--test scenarios obtained using the PSCC method and \ours on concatenated representations.
Specifically, for \ours we stack together the features from the 21st layer of CLIP ViT-L/14 and the features from 3rd layer of CLIP ResNet-50.
The representations extracted from ResNet-50 are bilinearly upsampled from $14 \times 14 \times D$ to $16 \times 16 \times D$
to match the spatial resolution of the features extracted by ViT-L/14; here $D$ denotes the feature dimension.
The representations from both networks have the same dimension $D = 1024$.
By concatenating the features along the last axis, we obtain a block of size $16 \times 16 \times 2048$, which then fed as input to the conv-20 decoder.

Compared to PSCC, \ours shows better generalization capabilities (results in the out-of-domain setups, off-principal diagonal), especially when trained on LDM and P2.
PSSC generally has better in-domain performance (principal diagonal), with the exception of the harder LDM--LDM case,
where \ours performs better (51.1\% compared to 41.5\% IoU)

\section{Additional qualitative results on Dolos}
\begin{figure*}[htb!]
\vspace{1.5cm}
\includegraphics[trim={0 0.2cm 0 0.2cm},clip,width=\textwidth]{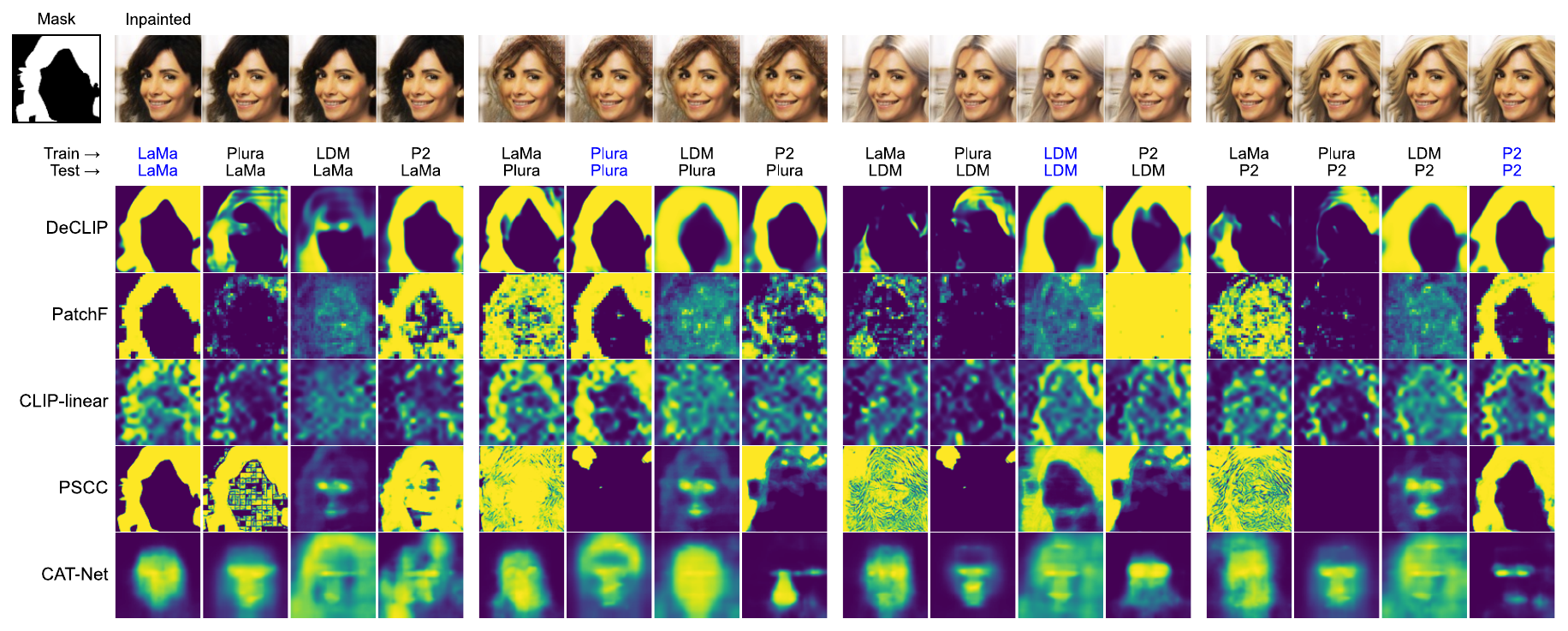}

\includegraphics[trim={0 0.2cm 0 0.2cm},clip,width=\textwidth]{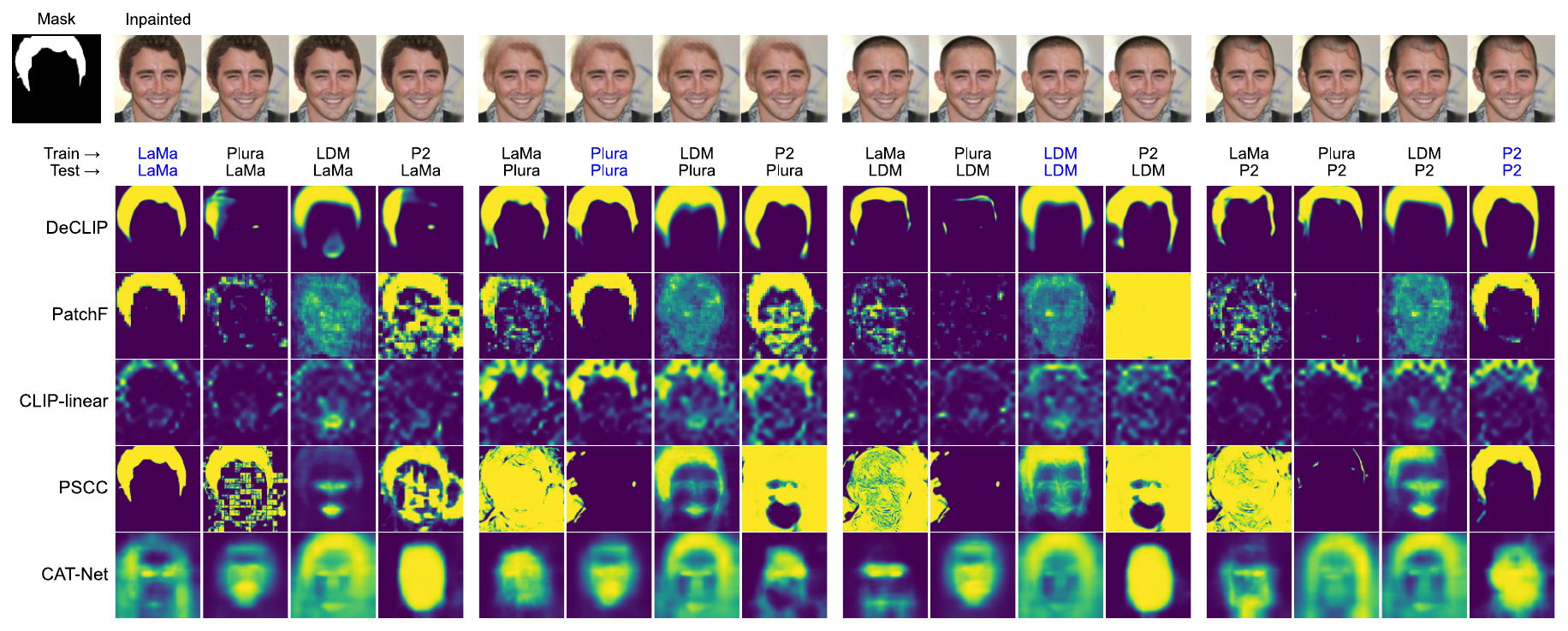}

\includegraphics[trim={0 0.2cm 0 0.2cm},clip,width=\textwidth]{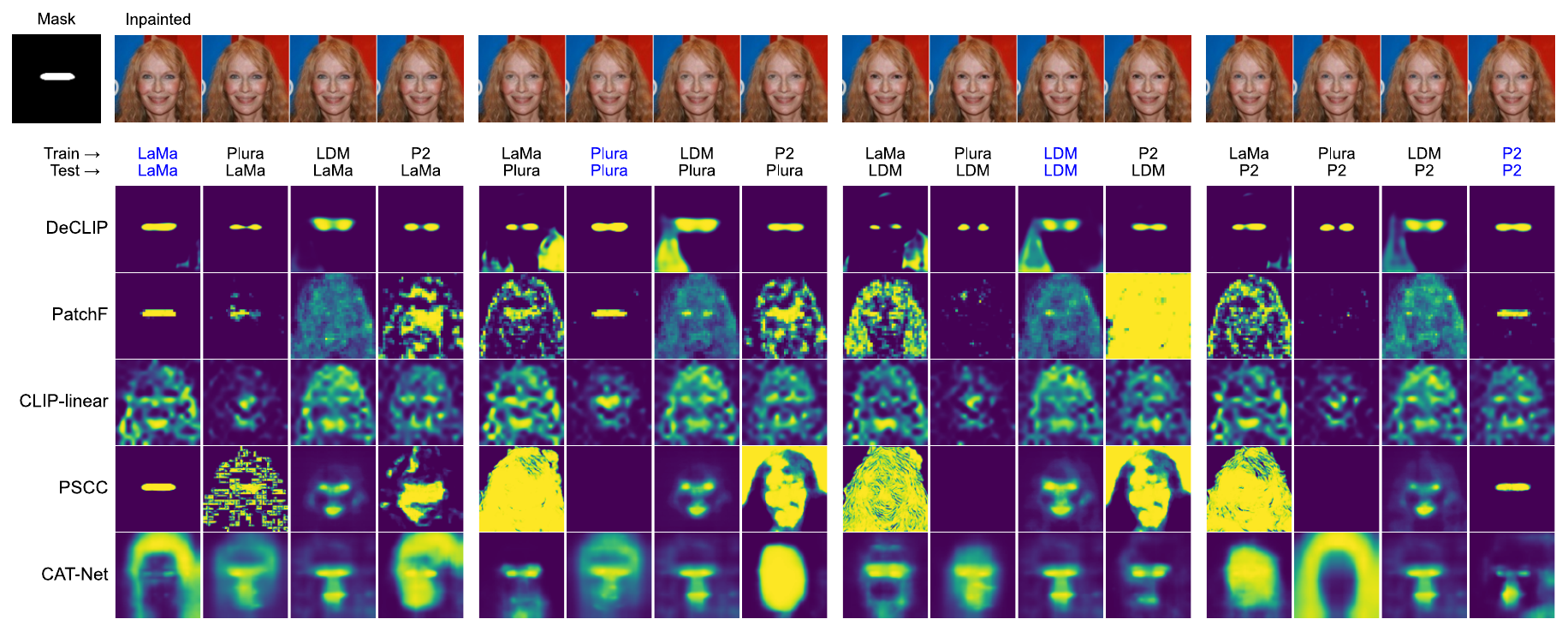}

\caption{%
Sample predictions for \ours (second row) and four other methods (Patch Forensics, CLIP-linear, PSCC, CAT-Net) on all 16 train--test combinations from the Dolos dataset.
The in-domain combinations are highlighted in blue; the others are out-of-domain combinations.
The black-and-white image in the top left corner shows the inpainting mask (white is the inpainted region) and the rest of the images in the first row are the inpainted images with one of the four test datasets (LaMa, Pluralistic, LDM, P2).
}
\label{fig:qualitative_1}
\end{figure*}

\begin{figure*}[htb!]

\includegraphics[trim={0 0.2cm 0 0.2cm},clip,width=\textwidth]{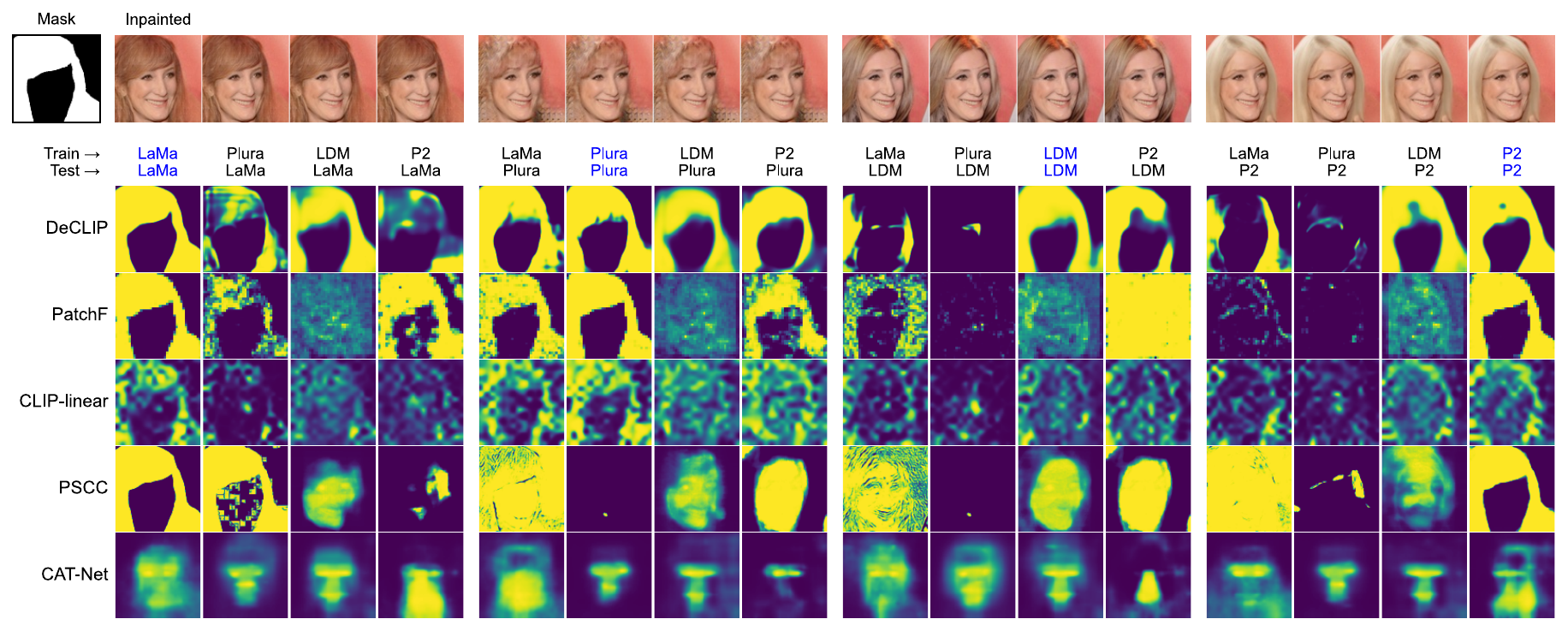}
\includegraphics[trim={0 0.2cm 0 0.2cm},clip,width=\textwidth]{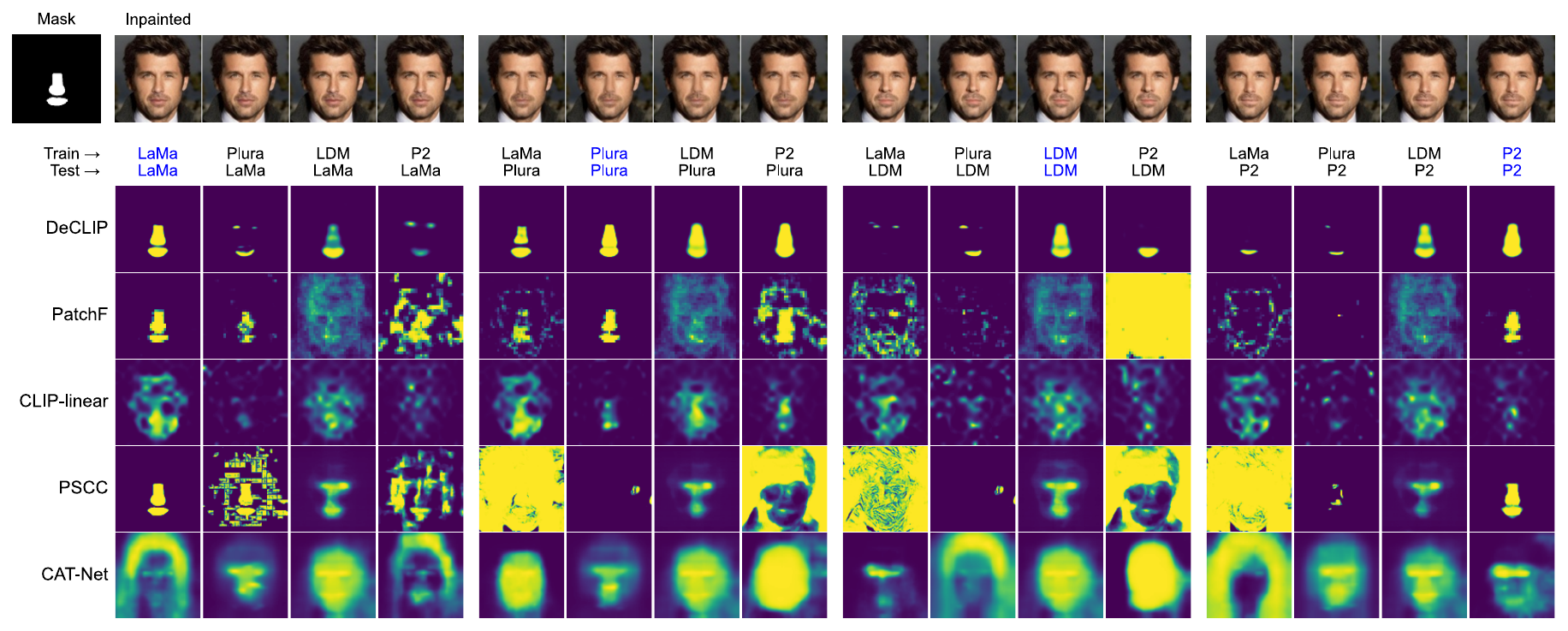}
\includegraphics[trim={0 0.2cm 0 0.2cm},clip,width=\textwidth]{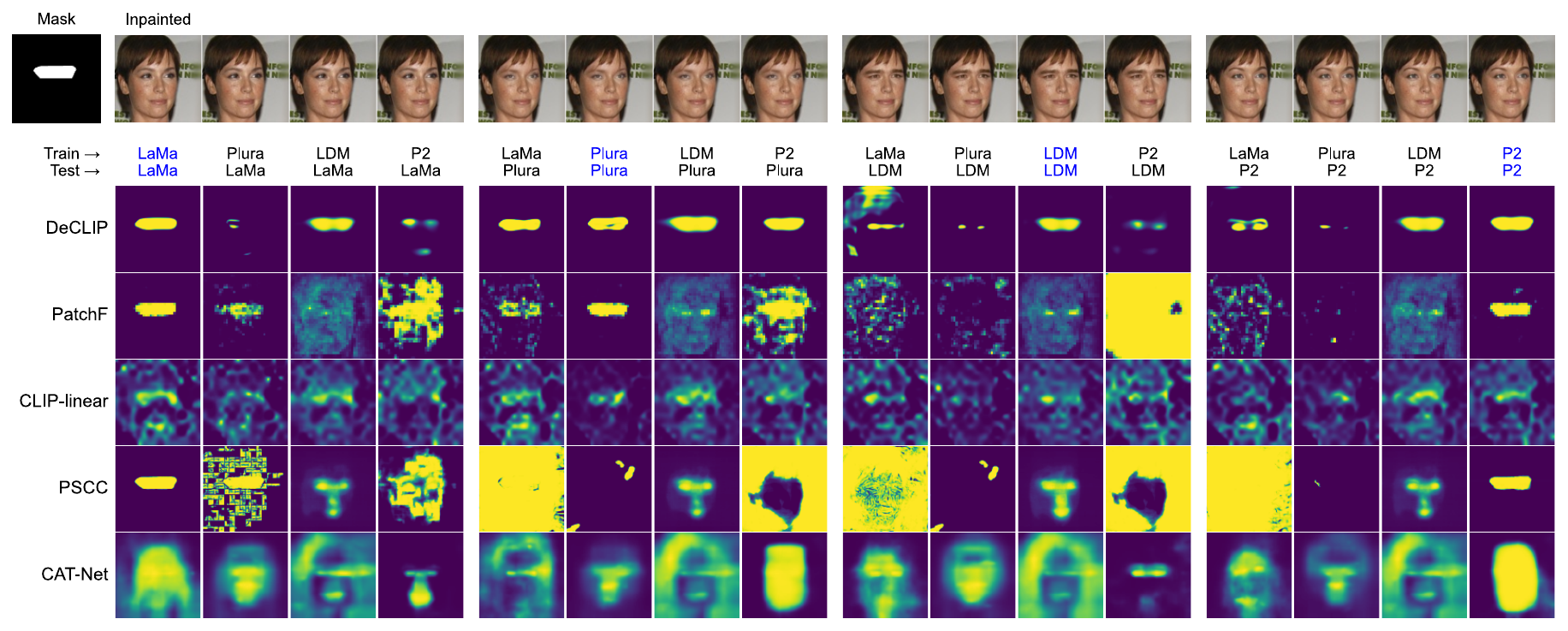}

\caption{%
Sample predictions for \ours (second row) and four other methods (Patch Forensics, CLIP-linear, PSCC, CAT-Net) on all 16 train--test combinations from the Dolos dataset.
The in-domain combinations are highlighted in blue; the others are out-of-domain combinations.
The black-and-white image in the top left corner shows the inpainting mask (white is the inpainted region) and the rest of the images in the first row are the inpainted images with one of the four test datasets (LaMa, Pluralistic, LDM, P2).
}
\label{fig:qualitative_2}
\end{figure*}

\begin{figure*}[htb!]

\includegraphics[trim={0 0.2cm 0 0.2cm},clip,width=\textwidth]{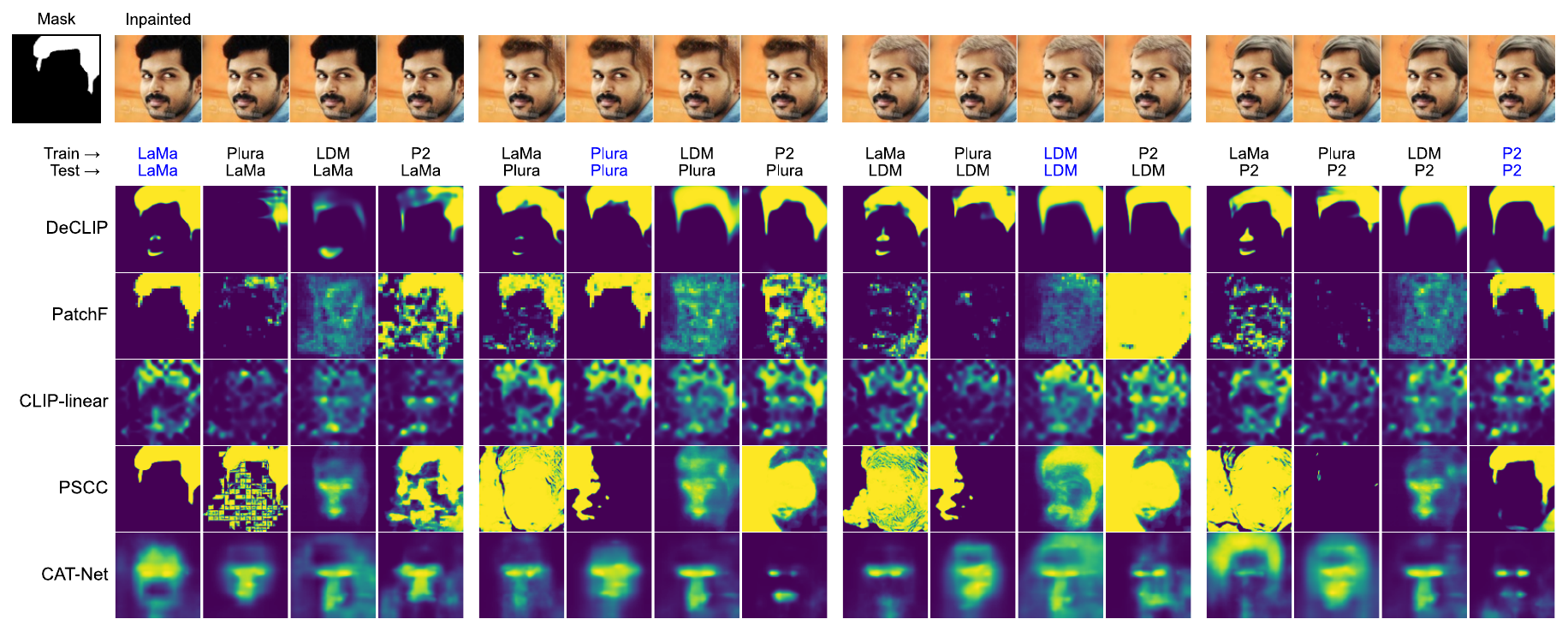}
\includegraphics[trim={0 0.2cm 0 0.2cm},clip,width=\textwidth]{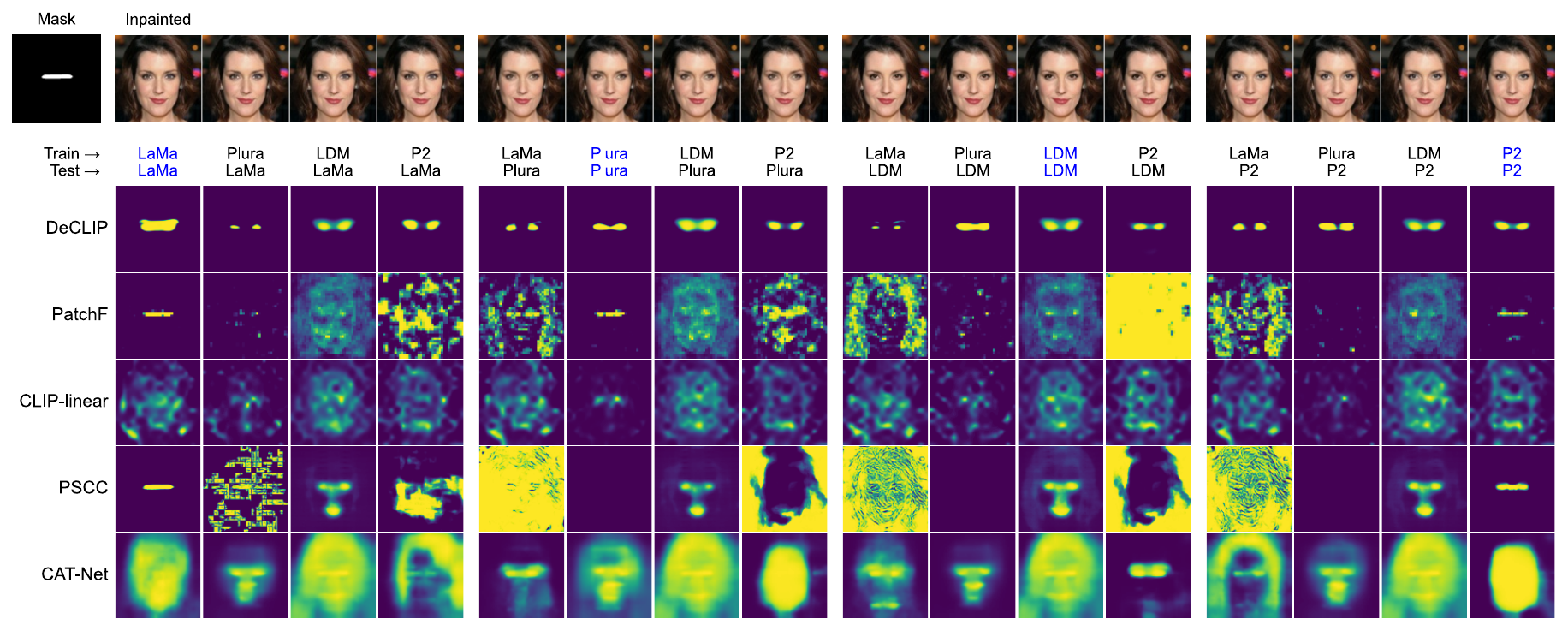}
\includegraphics[trim={0 0.2cm 0 0.2cm},clip,width=\textwidth]{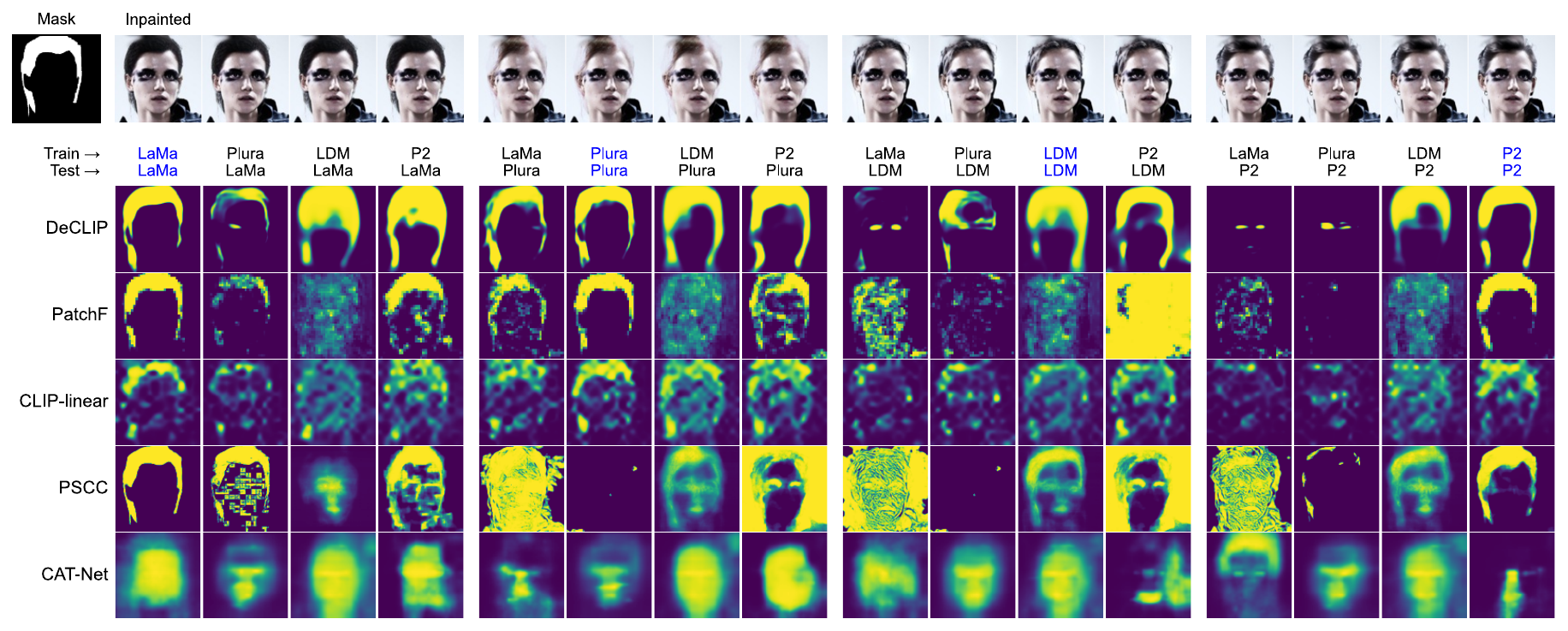}

\caption{%
Sample predictions for \ours (second row) and four other methods (Patch Forensics, CLIP-linear, PSCC, CAT-Net) on all 16 train--test combinations from the Dolos dataset.
The in-domain combinations are highlighted in blue; the others are out-of-domain combinations.
The black-and-white image in the top left corner shows the inpainting mask (white is the inpainted region) and the rest of the images in the first row are the inpainted images with one of the four test datasets (LaMa, Pluralistic, LDM, P2).
}
\label{fig:qualitative_3}
\end{figure*}

\begin{figure*}[htb!]

\includegraphics[trim={0 0.2cm 0 0.2cm},clip,width=\textwidth]{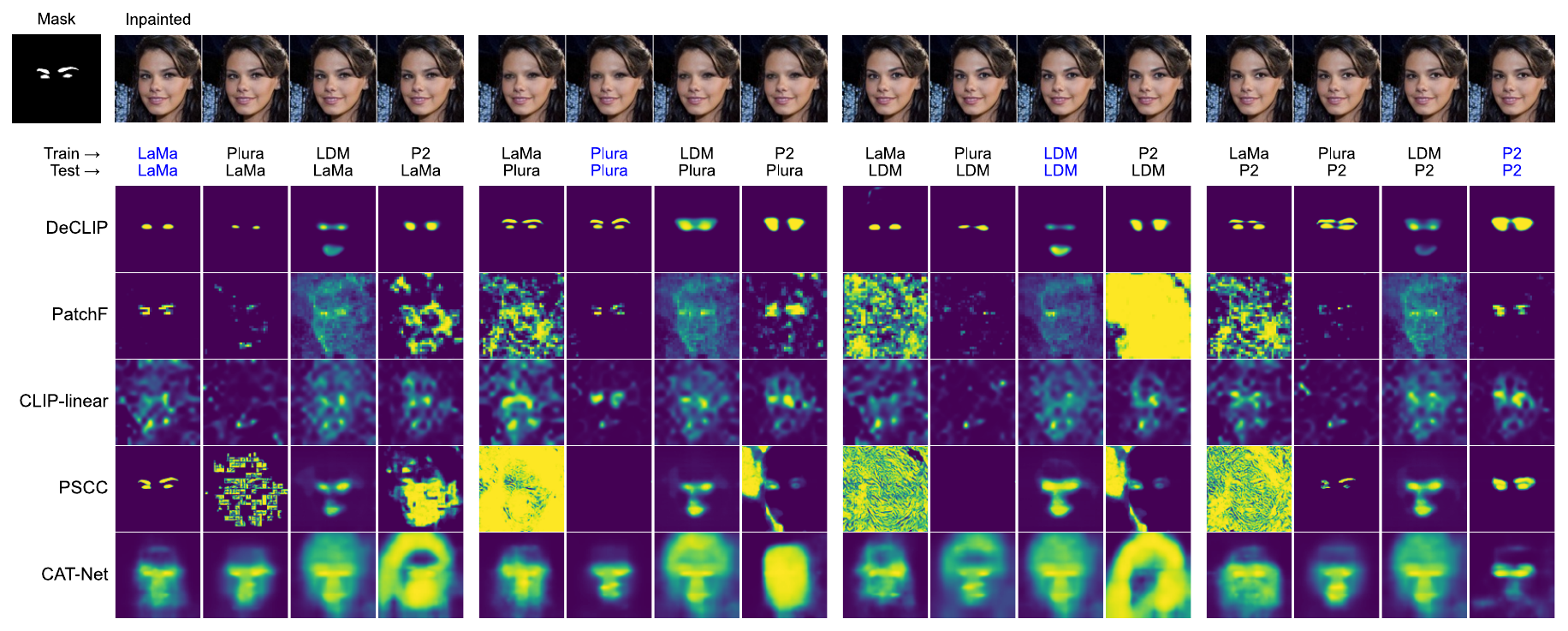}
\includegraphics[trim={0 0.2cm 0 0.2cm},clip,width=\textwidth]{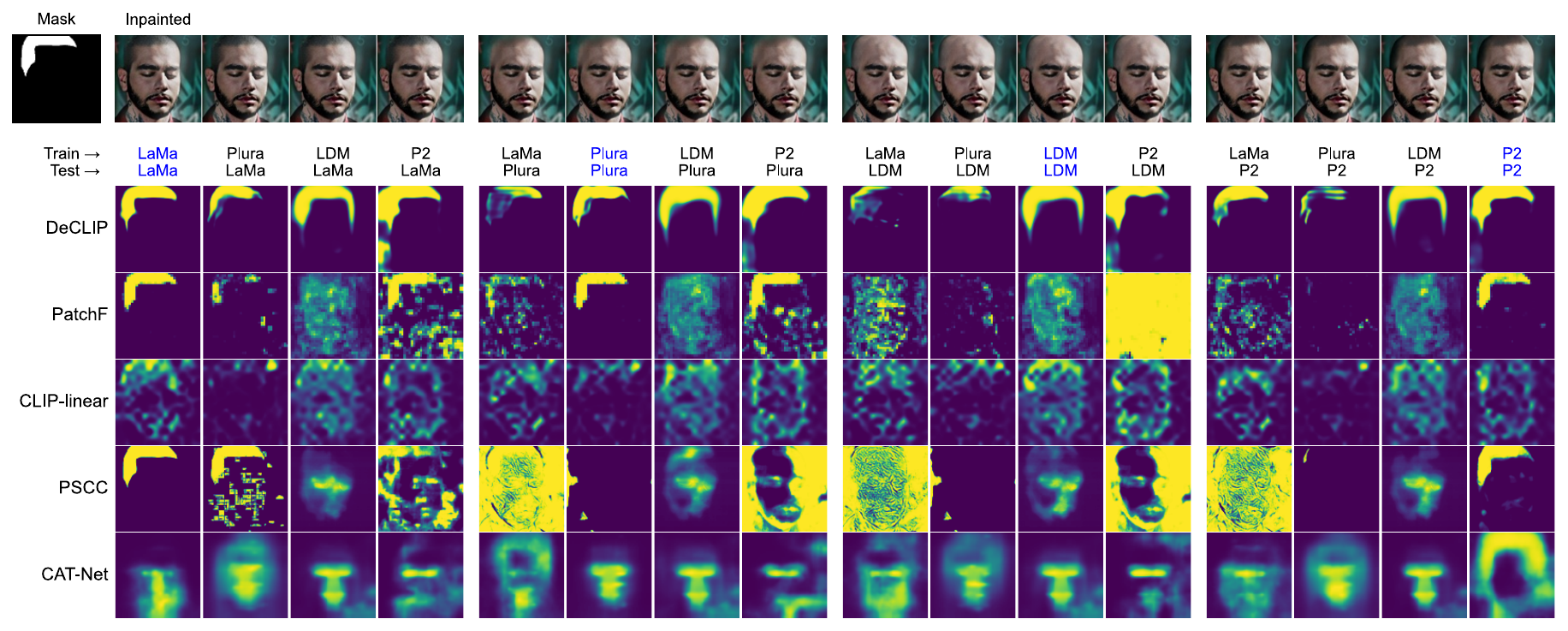}

\caption{%
Sample predictions for \ours (second row) and four other methods (Patch Forensics, CLIP-linear, PSCC, CAT-Net) on all 16 train--test combinations from the Dolos dataset.
The in-domain combinations are highlighted in blue; the others are out-of-domain combinations.
The black-and-white image in the top left corner shows the inpainting mask (white is the inpainted region) and the rest of the images in the first row are the inpainted images with one of the four test datasets (LaMa, Pluralistic, LDM, P2).
}
\label{fig:qualitative_4}
\end{figure*}

In Figures \ref{fig:qualitative_1}, \ref{fig:qualitative_2}, \ref{fig:qualitative_3}, \ref{fig:qualitative_4} we show detailed visual results on Dolos dataset for all train--test scenarios for \ours as well as four other methods trained and tested in the same way: Patch Forensics, CLIP-linear, PSCC and CAT-Net.
The results show that although some train--test scenarios are considerably harder than the other, \ours offers a plausible manipulation mask in the majority of cases.
We showcase different types of masks, from the very small ones that cover only eyes to larger ones that correspond to face and hair.
Patch Forensics and PSCC usually work well in domain (with the exception of LDM--LDM scenario), but genrally struggle in the out-of-domain cases.
CLIP-linear and CAT-Net struggle both in domain and out of domain, producing masks with arbitrary activations that follow the face characteristics.

\section{Additional qualitative results on COCO-SD}
\begin{figure}
    \centering
   \includegraphics[clip,width=0.48\textwidth]{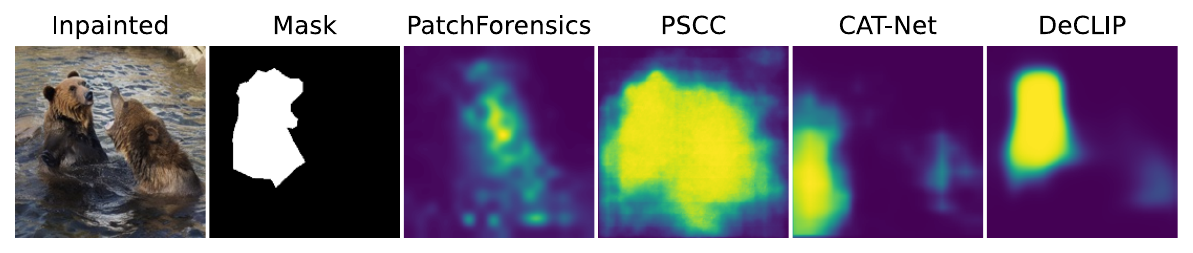}
     \includegraphics[clip,width=0.48\textwidth]{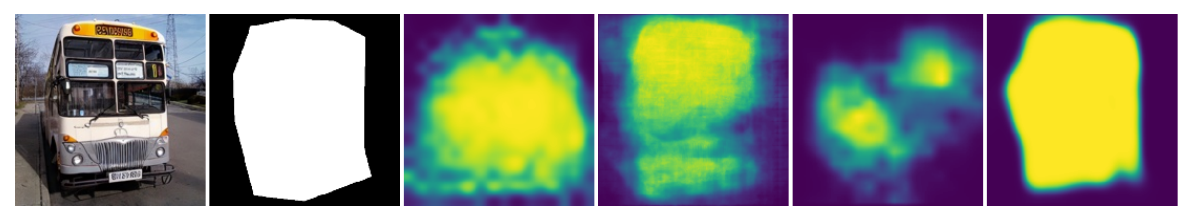}
      \includegraphics[clip,width=0.48\textwidth]{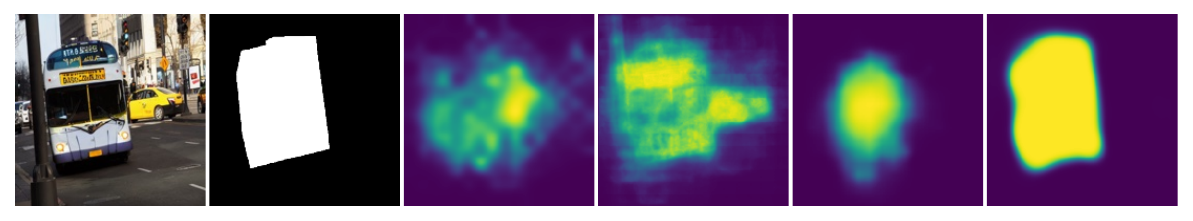}

      \includegraphics[clip,width=0.48\textwidth]{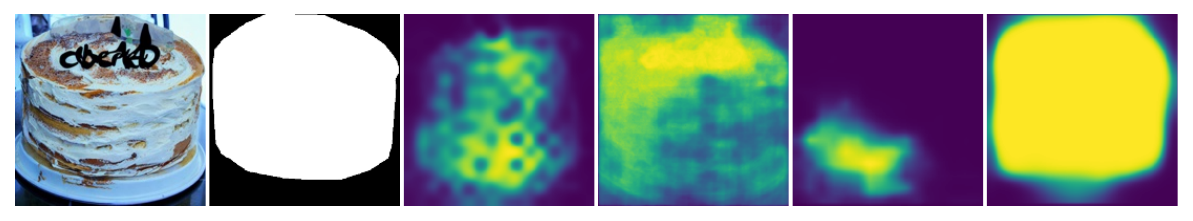}
      \includegraphics[clip,width=0.48\textwidth]{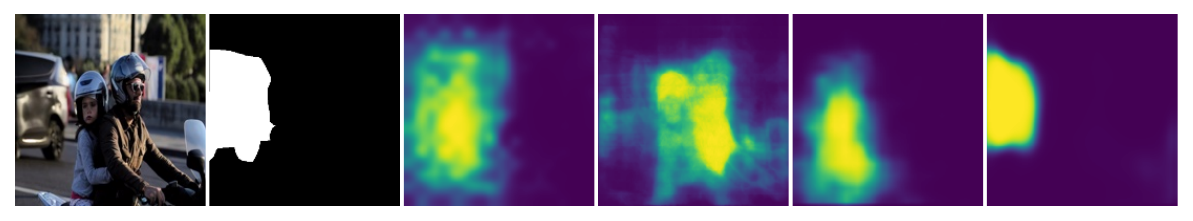}
      \includegraphics[clip,width=0.48\textwidth]{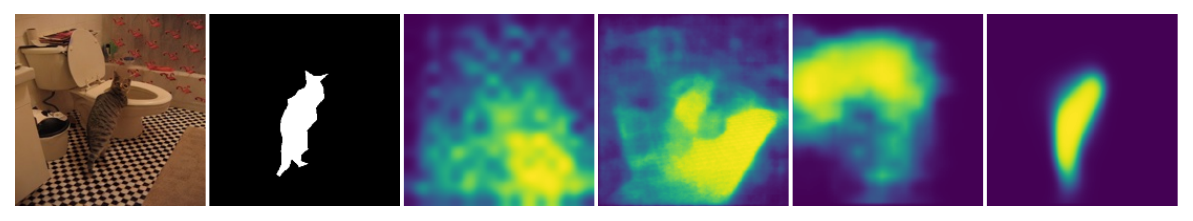}
      \includegraphics[clip,width=0.48\textwidth]{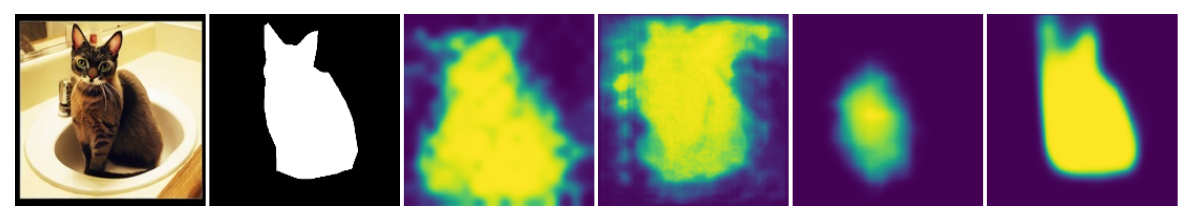}
      \includegraphics[clip,width=0.48\textwidth]{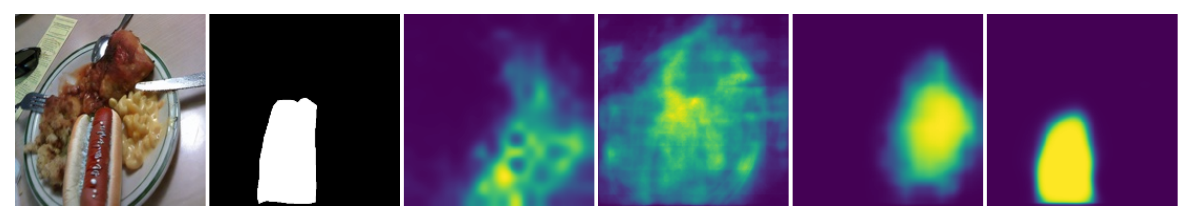}
      \includegraphics[clip,width=0.48\textwidth]{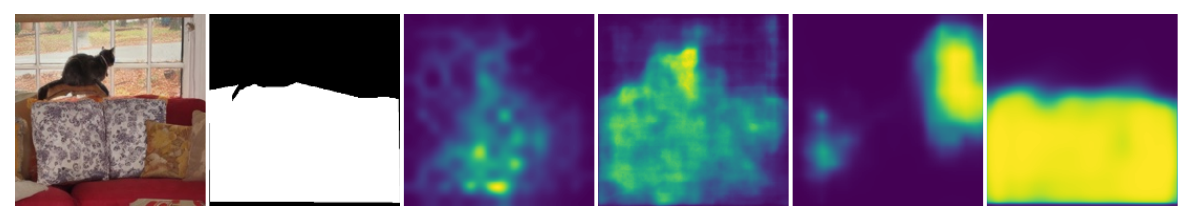}
       \includegraphics[trim={0 0 0 0.0cm},clip,width=0.48\textwidth]{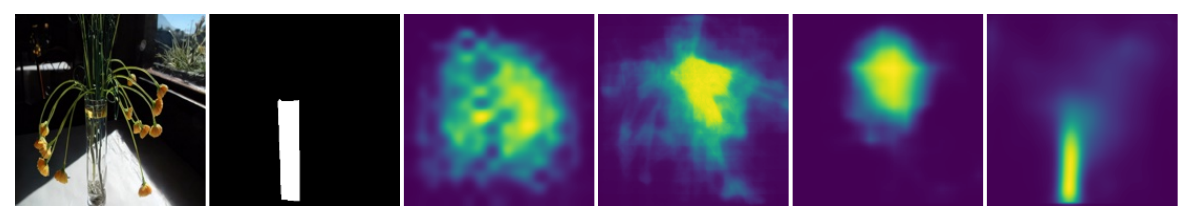}
    \includegraphics[trim={0 0 0 0.0cm},clip,width=0.48\textwidth]{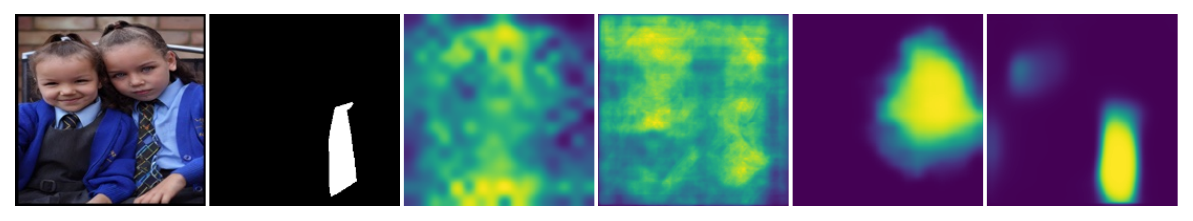}
    \caption{Manipulation localization results on COCO-SD, which has a more challenging set of masks and diverse content. \ours offers a more precise localization of the manipulated area.}
    \label{fig:cocosd_1}
\end{figure}

\begin{figure}
    \centering
     \includegraphics[trim={0 0 0 0.0cm},clip,width=0.48\textwidth]{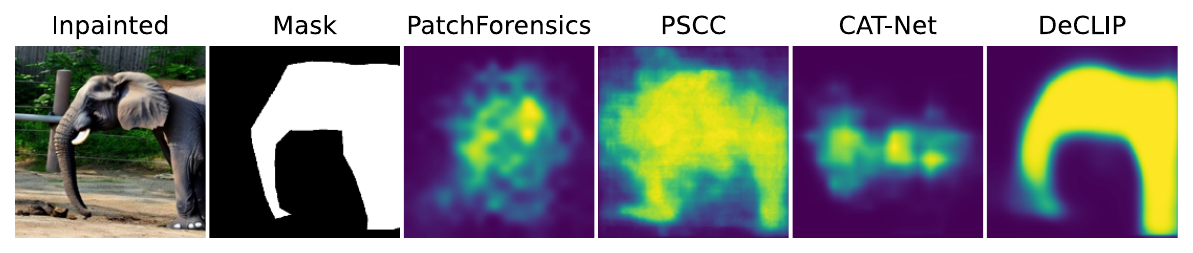}
    \includegraphics[trim={0 0 0 0.0cm},clip,width=0.48\textwidth]{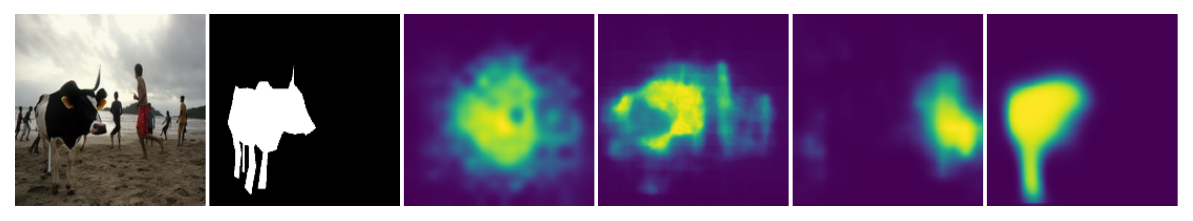} 
    \includegraphics[trim={0 0 0 0.0cm},clip,width=0.48\textwidth]{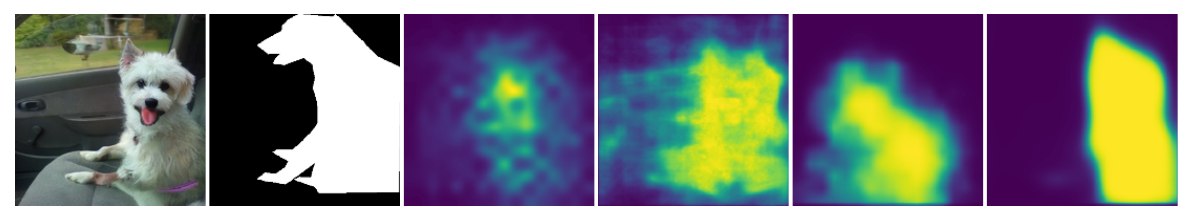}
    \includegraphics[trim={0 0 0 0.0cm},clip,width=0.48\textwidth]{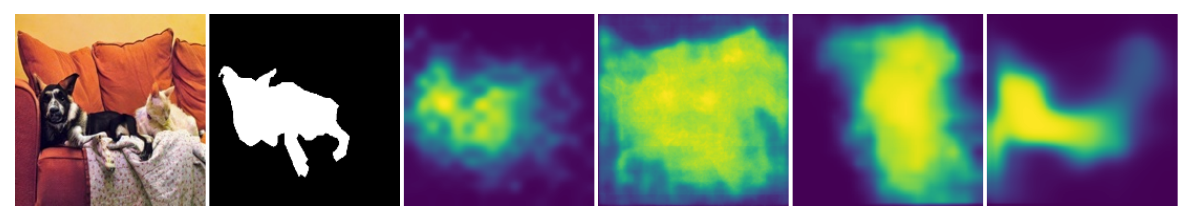}
    \includegraphics[trim={0 0 0 0.0cm},clip,width=0.48\textwidth]{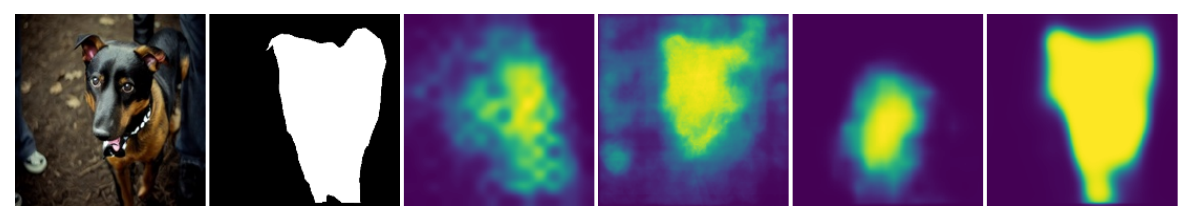}
    \includegraphics[trim={0 0 0 0.0cm},clip,width=0.48\textwidth]{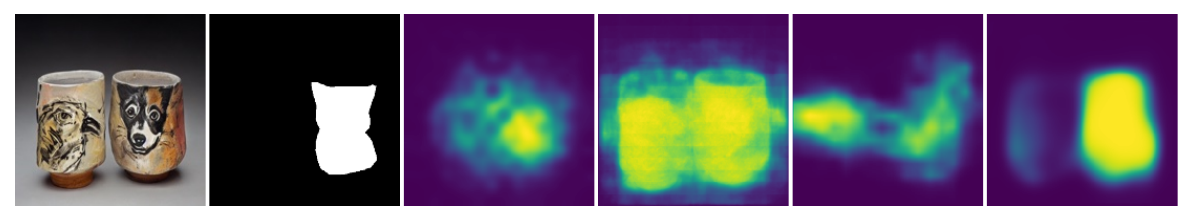}
    \includegraphics[trim={0 0 0 0.0cm},clip,width=0.48\textwidth]{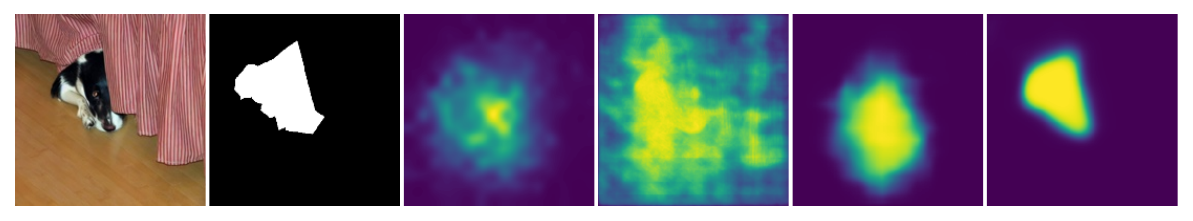}
    \includegraphics[trim={0 0 0 0.0cm},clip,width=0.48\textwidth]{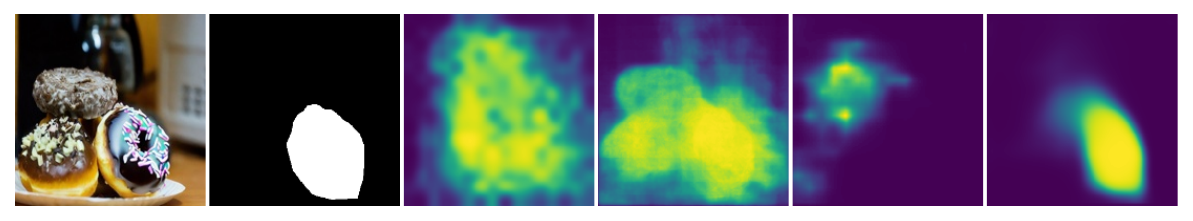}
    \includegraphics[trim={0 0 0 0.0cm},clip,width=0.48\textwidth]{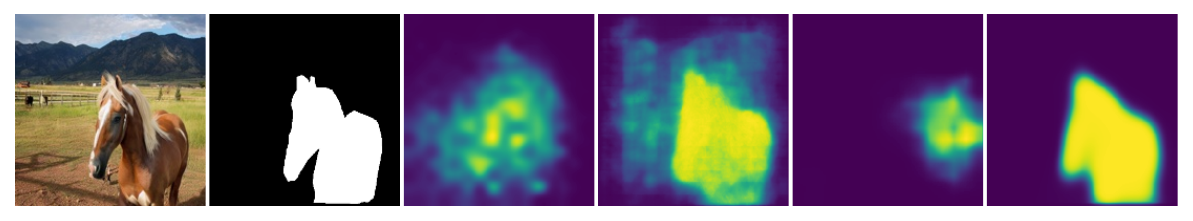}
    \includegraphics[trim={0 0 0 0.0cm},clip,width=0.48\textwidth]{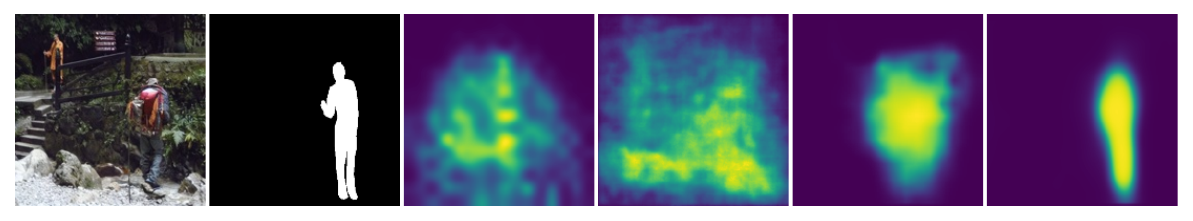}
    \includegraphics[trim={0 0 0 0.0cm},clip,width=0.48\textwidth]{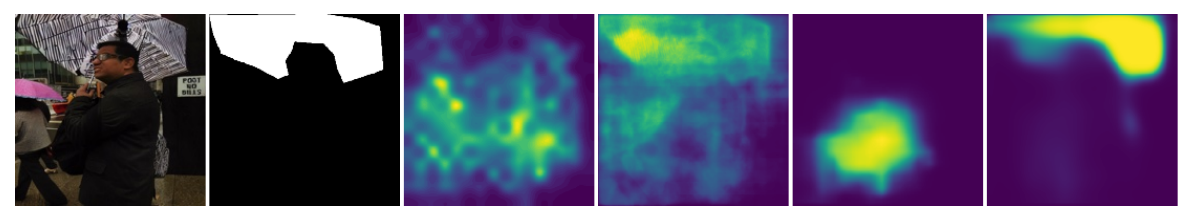}

    \caption{Manipulation localization results on COCO-SD, which has a more challenging set of masks and diverse content. \ours offers a more precise localization of the manipulated area.}
    \label{fig:cocosd_2}
\end{figure}

In Figures \ref{fig:cocosd_1} and \ref{fig:cocosd_2} we provide additional results on COCO-SD dataset for \ours, Patch Foreniscs, PSCC and CAT-Net.
Notice that even in a diverse visual domain, with arbitrary-shaped inpainted regions, \ours has a more stable and precise localization of the manipulated area.
The dataset is particularly hard as the inpainted objects are often parts of a larger one (e.g. the tie, the drawing of the dog on a cup), represent a single entity among similar of the same type (the doughnut, the bowl).
Even in these conditions, \ours provides plausible maps of the inpainting.

\section{Illustration of LDM images}

\begin{figure}[htb!]
    \includegraphics[trim={0 0.2cm 0 0.1cm},clip,width=0.94\linewidth]{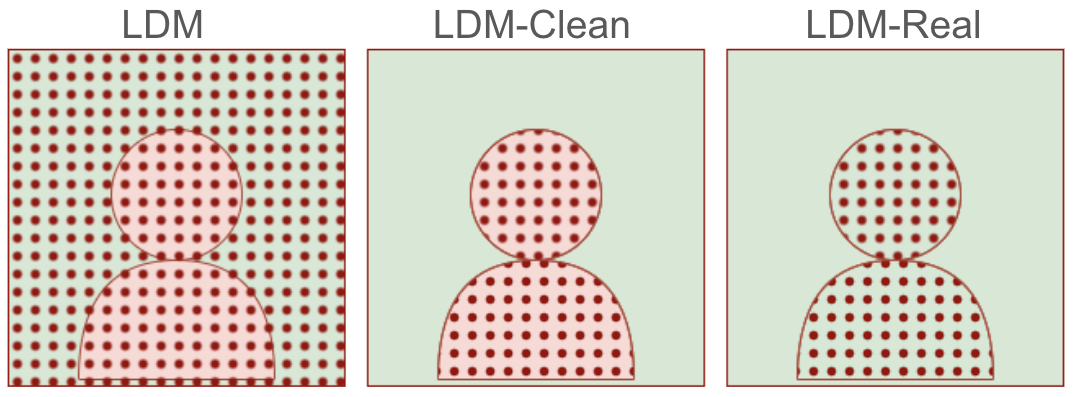}
    \caption{%
    Schematic view of different types of inpaintings with LDM considered in Section 5, Table 4 in the main paper.  
    }
    \label{fig: ldm_types}
\end{figure}

In Figure~\ref{fig: ldm_types} we show how fingerprint and fake content are distributed in different types of LDM images. Green color corresponds to real content, red color corresponds to fake content and red dots symbolize fingerprint. 

{\small
\bibliographystyle{ieee_fullname}
}